\documentclass[conference]{IEEEtran}
\IEEEoverridecommandlockouts
\usepackage{cite}
\usepackage{amsmath,amssymb,amsfonts}
\usepackage{algorithm}
\usepackage{algorithmic}
\usepackage{graphicx}
\usepackage{textcomp}
\usepackage{xcolor}
\newcommand{\eat}[1]{}
\usepackage{booktabs}
\usepackage{subfigure}
\usepackage{multirow}
\usepackage{hyperref}
\newtheorem{myDef}{Definition}

\def\BibTeX{{\rm B\kern-.05em{\sc i\kern-.025em b}\kern-.08em
    T\kern-.1667em\lower.7ex\hbox{E}\kern-.125emX}}
\newcommand\btrevise[1]{\textcolor{black}{#1}}

\begin{document}

\title{Decision Support System for Chronic Diseases Based on Drug-Drug Interactions}

\author{
Tian Bian\textsuperscript{\rm 1}, 
Yuli Jiang\textsuperscript{\rm 1},
Jia Li\textsuperscript{\rm 2*}\thanks{$^*$Corresponding authors.},
Tingyang Xu\textsuperscript{\rm 3*},
Yu Rong\textsuperscript{\rm 3},\\
Yi Su\textsuperscript{\rm 4},
Timothy Kwok\textsuperscript{\rm 1},
Helen Meng\textsuperscript{\rm 1},
Hong Cheng\textsuperscript{\rm 1}\\
% \thanks{corresponding author}\\ % All authors must be in the same font size and format. Use \Large and 
% \textbf to achieve this result when breaking a line
% \textsuperscript{\rm 1}  Department of Systems Engineering and Engineering Management, The Chinese University of Hong Kong\\ 
% \textsuperscript{\rm 2}Data Science and Analytics Thrust, Hong Kong University of Science and Technology\\
% \textsuperscript{\rm 3}Tencent AI Lab\\ 
% \textsuperscript{\rm 4}School of Medicine,  Hunan Normal University\\
% \textsuperscript{\rm 5}Department of Medicine and Therapeutics, The Chinese University of Hong Kong\\
\textsuperscript{\rm 1}  The Chinese University of Hong Kong,\\
\textsuperscript{\rm 2} Hong Kong University of Science and Technology (Guangzhou),\\
\textsuperscript{\rm 3} Tencent AI Lab, 
\textsuperscript{\rm 4} Hunan Normal University\\
\{tbian,hmmeng,hcheng\}@se.cuhk.edu.hk, yljiang@cse.cuhk.edu.hk, jialee@ust.hk, \\ tingyangxu@tencent.com, yu.rong@hotmail.com, alddle@hunnu.edu.cn, tkwok@cuhk.edu.hk\\
% \{tingyangxu, masonzhao, joehhuang\}@tencent.com \\
}

\maketitle

\begin{abstract}
Many patients with chronic diseases resort to multiple medications to relieve various symptoms, which raises concerns about the safety of multiple medication use, as severe drug-drug antagonism can lead to serious adverse effects or even death.
This paper presents 
% \rr{If this is an industrial track paper, I think `platform' is more suitable.} 
a Decision Support System, called DSSDDI, based on drug-drug interactions to support doctors prescribing decisions. 
DSSDDI contains three modules, Drug-Drug Interaction (DDI) module, Medical Decision (MD) module and Medical Support (MS) module.
%By avoiding the antagonistic effects between drugs and promoting their synergistic effects, 
The DDI module learns safer and more effective drug representations from the drug-drug interactions.
% \rr{This an plain description. Why we need counterfactual links here?}
% DSSDDI learns the drug relation representations through the proposed drug-drug interaction (DDI) module. It not only avoids the antagonistic effects between the suggested drugs, but also considers the synergistic effects between the drugs, thus providing safer and more efficient treatment regimens. 
To capture the potential causal relationship between DDI and medication use, the MD module considers the representations of patients and drugs as context, DDI and patients' similarity as treatment, and medication use as outcome to construct counterfactual links for the representation learning.
% \rr{`the causal representation learning' is not good.}
%On the other hand, with the proposed medication suggestion module, DSSDDI realizes personalized medication suggestions which improves the relevance of the medication regimen to the patient's features. 
% \rrrevise{
% Furthermore, we propose the MS module to provide the suggesting drug candidates to doctor with explanations.
% }
Furthermore, the MS module provides drug candidates to doctors with explanations.
Experiments on the chronic data collected from the Hong Kong Chronic Disease Study Project \btrevise{and a public diagnostic data MIMIC-III}
demonstrate that DSSDDI can be a reliable reference for doctors in terms of safety and efficiency of clinical diagnosis, with significant improvements compared to baseline methods. Source code of the proposed DSSDDI is publicly available at \url{https://github.com/TianBian95/DSSDDI}.
\end{abstract}


\section{Introduction}
Due to physiological changes, increased risk of disease and decreased drug clearance, problematic polypharmacy has become a significant factor in the increased risk of severe Adverse Drug Events (ADEs), hospital admissions, and death in chronic patients \cite{wimmer2017clinical}.  This issue is especially prominent during critical times, such as the epidemic of coronavirus disease (COVID-19). With the shortage of clinical resources, medication for chronic patients lacks guidance from professional doctors and presents unique challenges.
Further, polypharmacy increases the potential for drug-drug interactions (DDI), including potentially inappropriate drug combinations present in prescription medications \cite{dumbreck2015drug}, which accelerates the imbalance between the complex needs of the chronic patients and the problems caused by the multiple medications. A systematic approach is required to efficiently support doctors in tailoring of medication regimens to extricate the chronic patients from the dilemma. 

Advanced technologies nowadays have been applied to develop more effective decision support systems for better-informed decisions \cite{awaysheh2019review}. 
However, some methods~\cite{DBLP:conf/icde/JinZ00W20, DBLP:conf/sigir/ZhangYZC021} that learn from association between patients and drugs, mainly rely on patient and drug features, but ignore the impact of DDI on medical decisions. 
Some other methods~\cite{DBLP:conf/ijcai/YangXMGS21} make use of DDI to learn drug embeddings but fail to capture the causal relationship between drug embeddings and medication suggestions.
% Once a system was designed to strictly follow the DDI relations, it may miss some drug combinations that have weak interactions yet are effective for the patients' symptoms.
Therefore, our goal in this paper is: \textbf{(1) leveraging DDI to avoid severe ADEs in medication suggestions, and (2) employing the causal model \cite{hu2021distilling,yang2021deconfounded} to learn the potential causal relationships between DDI and medication suggestions to improve the accuracy}.

% Another important consideration is that the proposed decision support system is mainly for patients suffering from chronic diseases. So we need to make sure that the training data is representative of such groups with chronic diseases, and at the same time ensure that candidate drugs are commonly-taken drugs by the group. To this end, 
%Hong Kong Chronic Disease Study project正在同医院合作部署AI models以辅助慢性病的预测和治疗,因此DSSDDI被用于向医生提供更安全和高效的处方参考。

\begin{figure}[t]
  \centering
  \includegraphics[width=0.62\linewidth]{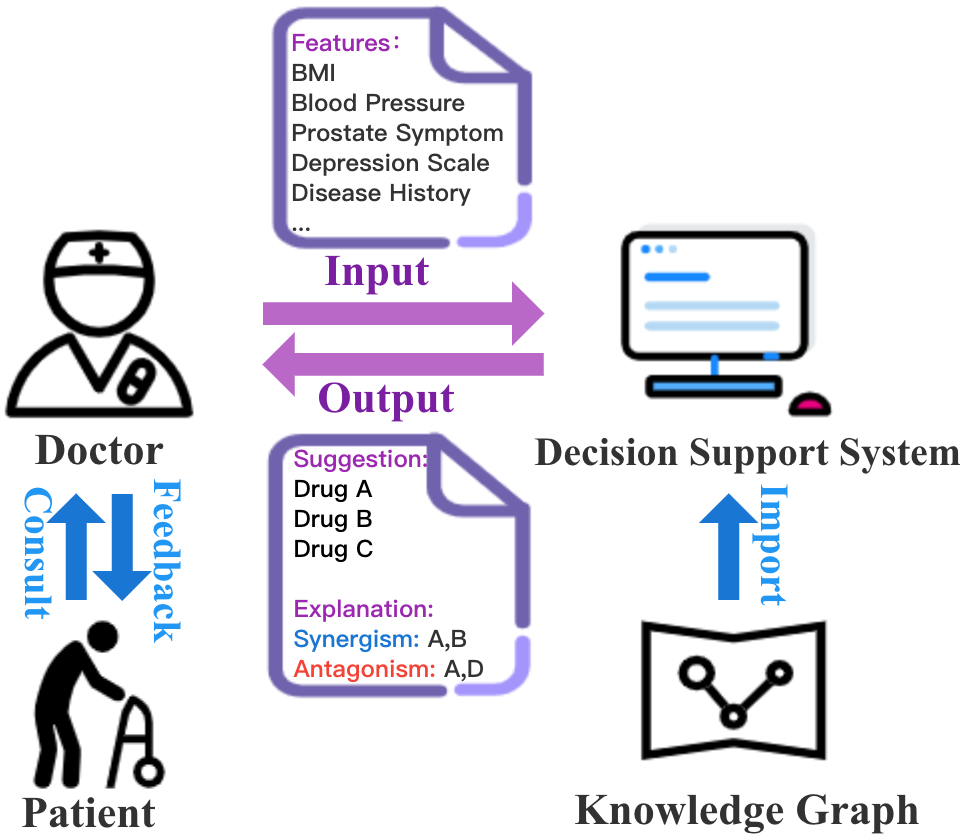}
  \vspace{-0.3cm}
  \caption{Our proposed decision support system uses external knowledge of DDI. Given patient features as input, a doctor can obtain the medication suggestions from the system, as well as the corresponding DDI explanations.}
  \vspace{-0.4cm}
  \label{flowchart}
\end{figure}

We studied the chronic patients through the Hong Kong Chronic Disease Study Project, including their personal features, clinical history, psychological assessment and medication use.  We propose a decision support system, DSSDDI, that can provide explainable medication suggestions for chronic diseases.  A bipartite graph can be naturally formed on the patients and drugs, and then DSSDDI applies link prediction on the bipartite graph for medication suggestions. With the help of an external DDI knowledge graph, the decision support system inputs patient features and outputs medication suggestions and the corresponding DDI explanations to doctors as clinical diagnostic references. The design of DSSDDI is depicted in Fig. \ref{flowchart}. 

\eat{DSSDDI is validated by industrial application.  is deploying AI models to aid in the treatment of chronic diseases. Through this project, we have validated the effectiveness of DSSDDI based on the population of chronic patients in Hong Kong. To make sure that the training data is representative of such groups with chronic diseases, we collect data from the Hong Kong Chronic Disease Study project in which subjects aged 65 years or over were recruited between 2001 - 2017. From this data set, we collect information on each subject's features and his/her medication use.} 

There are three modules in DSSDDI: the Drug-Drug Interaction (DDI) module, the Medical Decision (MD) module and the Medical Support (MS) module.
% a decision support system called \textbf{D}ecision \textbf{S}upport \textbf{S}ystem based on \textbf{D}rug-\textbf{D}rug \textbf{I}nteractions (DSSDDI).
% The main idea of DSSDDI is first to employ medical decision component that consists of DDI module and MS module to achieve better personalized medication suggestions by avoiding antagonistic effects among the suggested multiple drugs while maintaining synergistic effects.
The function and merit of each module are described as follows.

\begin{itemize}
    \item The DDI module uses our proposed Drug-Drug Interaction Graph Convolutional Network (DDIGCN) to learn drug representations from synergistic or antagonistic effects between drugs. The DDI module can alleviate severe adverse drug events, which is crucial to ensure safe and effective medical decisions.
    \item The MD module uses a Medical Decision Graph Convolutional Network (MDGCN) to suggest drugs. We construct counterfactual links to augment the training data for MDGCN based on the causal model that considers the representations of patients and drugs as context, DDI and patients' similarity as treatment, and medication use as outcome. This can  learn the causal relationship between DDI and medication use.

%    We design a novel MD module for the proposed decision support system that captures the potential causal relationship between DDIs and medication use, improving the effectiveness of medication suggestions based on the use of DDI knowledge.
    % \item The proposed decision support system develops a DDI module that suggests drugs with synergistic effects while avoiding inappropriate drugs for patients, which is important to ensure safe and effective medication decisions.
    % \item We design a novel decision support system that \btrevise{captures the causal relationship between DDIs and medication use through the proposed MD module, improving the effectiveness of medication suggestions based on the use of DDI knowledge.}
    % personalized medication suggestions through the proposed MS module, avoiding the the over-smoothing of patient representations due to the similar medication use of patients in the training data.
    \item After obtaining the suggested drugs, the MS module extracts coherent subgraphs with DDI knowledge as explainable factors for doctors' reference. Such subgraphs illustrate the synergistic and antagonistic effects between drugs.

%    We define Suggestion Satisfaction (SS) to measure the coherence of the extracted subgraphs. 

%    We also propose a MS module to analyze the synergistic and antagonistic interactions in the DDI graph and define the Suggestion Satisfaction (SS) measurement for the evaluation of the suggested drugs. 
    % \rr{How about the contribution of data collection and experimental justification?}
    % \item  
\end{itemize}
Experiments on data from Hong Kong Chronic Disease Study Project \btrevise{and public diagnostic data MIMIC-III~\cite{johnson2016mimic}} demonstrate the superior performance of DSSDDI in medication suggestion and its explainability.

% The rest of this paper is organized as follows. Section \ref{Profor} presents the problem formulation. Section \ref{ProDSSDDI} introduces the design of DSSDDI. Section \ref{datacol} describes our data collection procedure. Section \ref{exp} reports the experimental results. Section \ref{lessons} summarizes our experiences and insights gained from this study. Section \ref{review} reviews the related work. Finally, Section \ref{conclu} concludes the paper. \rr{We can remove this part if the space is insufficient. }

% \vspace{-0.1cm}
\section{Data Collection}\label{datacol}
Our study focuses on the patients participating in the Hong Kong Chronic Disease Study Project, who may require multiple medications because they have multi-chronic diseases. We collect data from questionnaire interviews and laboratory results to predict what medications they would need to take. Since drug-drug interactions are the primary consideration for doctors when prescribing medications, DSSDDI is designed to avoid the inclusion of drugs that contain antagonistic effects between each other and suggest drugs that have synergistic effects. In this section, we first introduce the participants enrolled in this project, then describe the drugs we use for the decision support system, followed by a further description of the drug-drug interactions used in this paper.

\subsection{Participants}
\begin{figure}[t]
  \centering
  \includegraphics[width=0.75\linewidth]{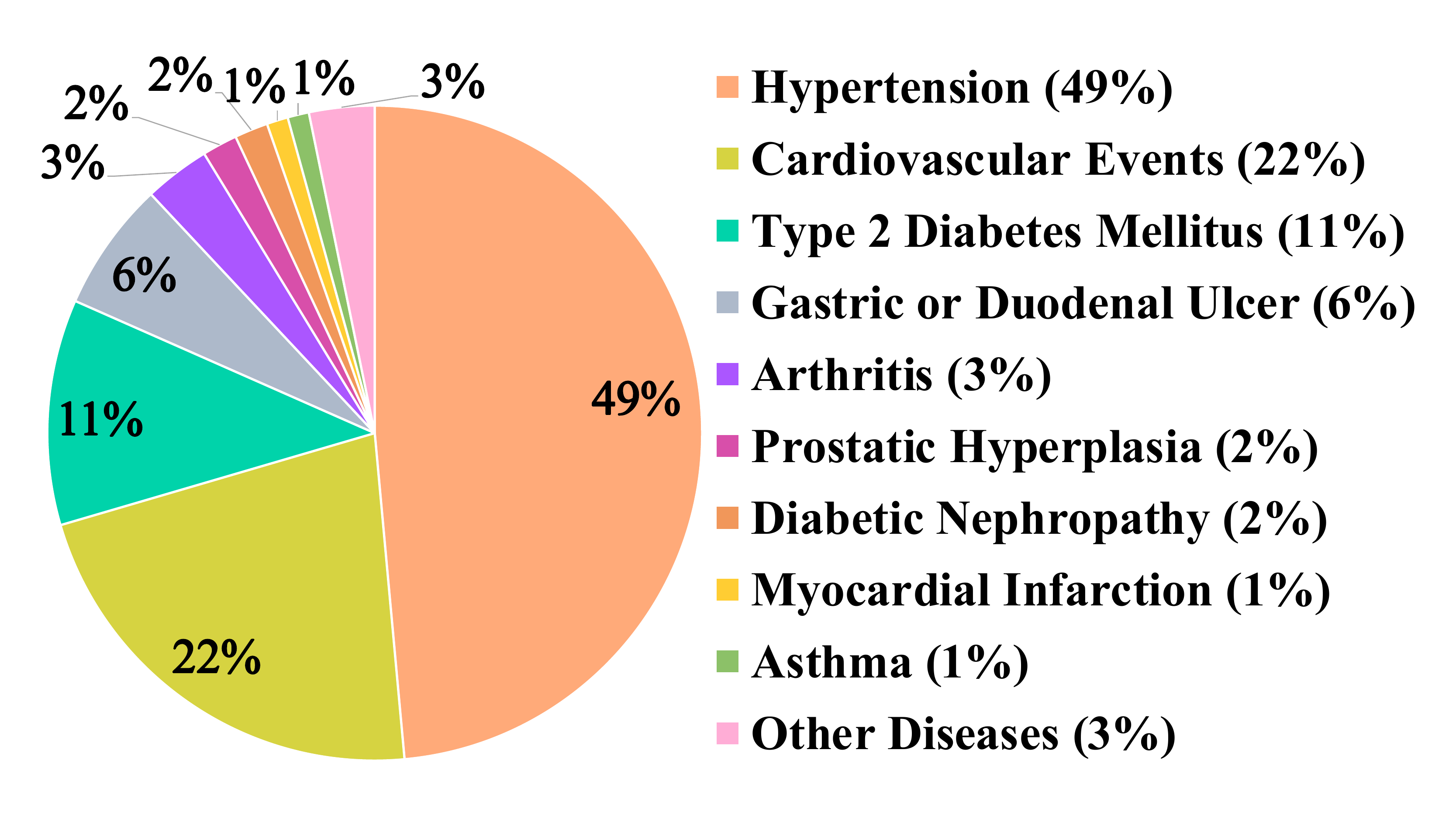}
  \vspace{-0.5cm}
  \caption{The proportion of patients with various diseases.}
  \vspace{-0.1cm}
  \label{drugPrec}
\end{figure}

This project was initiated by Prince of Wales Hospital\footnote{https://www3.ha.org.hk/pwh/index\_e.asp} in 2001. Subjects aged 65 years and older were recruited under the project. %During 2001 - 2017, 2000 Chinese men and 2000 Chinese women were enrolled.
% During 2001 - 2017, 2254 Chinese men and 1907 Chinese women were enrolled.
The cohort was invited for a questionnaire interview and measurement of physical performance for 1 -- 4 times during 2001 -- 2017.  %As the interval between two consecutive interviews of the same person is at least two years, we find their medications can be quite different, thus we treat each interview as one record.  Meanwhile, as we use the pretrained embeddings in Drug Repurposing Knowledge Graph (DRKG) \cite{drkg2020} as the original features of the drugs, some subjects who were taking drugs not in DRKG are ignored.  After the above data preprocessing process, 
We extracted 2254 male and 1903 female interview records. \eat{For brevity, we use `subject' instead of `record' in the following.}
The distribution of the diseases suffered by these subjects is shown in Fig. \ref{drugPrec}. Hypertension, cardiovascular diseases, diabetes, digestive diseases and arthritis are the chronic diseases they commonly suffer.

The questionnaire interview contains three types of subject information. The first type is the personal information about the participants such as gender and age. The second type is the clinical history of participants. For example, participants were asked whether they had prostatitis before or had taken taken the Alpha-blocker.  It is important to note that the questionnaire only mentions the clinical history of the family of drugs, but not the specific drugs. The third type is a psychological assessment of the participants, including the Geriatric Depression Scale (GDS) Score and some emotional questions, such as whether they had felt downhearted in the last four weeks. The physical examination included the participants' blood pressure, Body Mass Index (BMI), etc. Combining the three types of information, we collected a total of 71 features. % for each participant. 
% More specific descriptions of these features are listed in Table \ref{tab:subjectFeat}.

\vspace{-0.2cm}
\subsection{Medication Use}
\begin{figure}[t]
  \centering
  \includegraphics[width=0.8\linewidth]{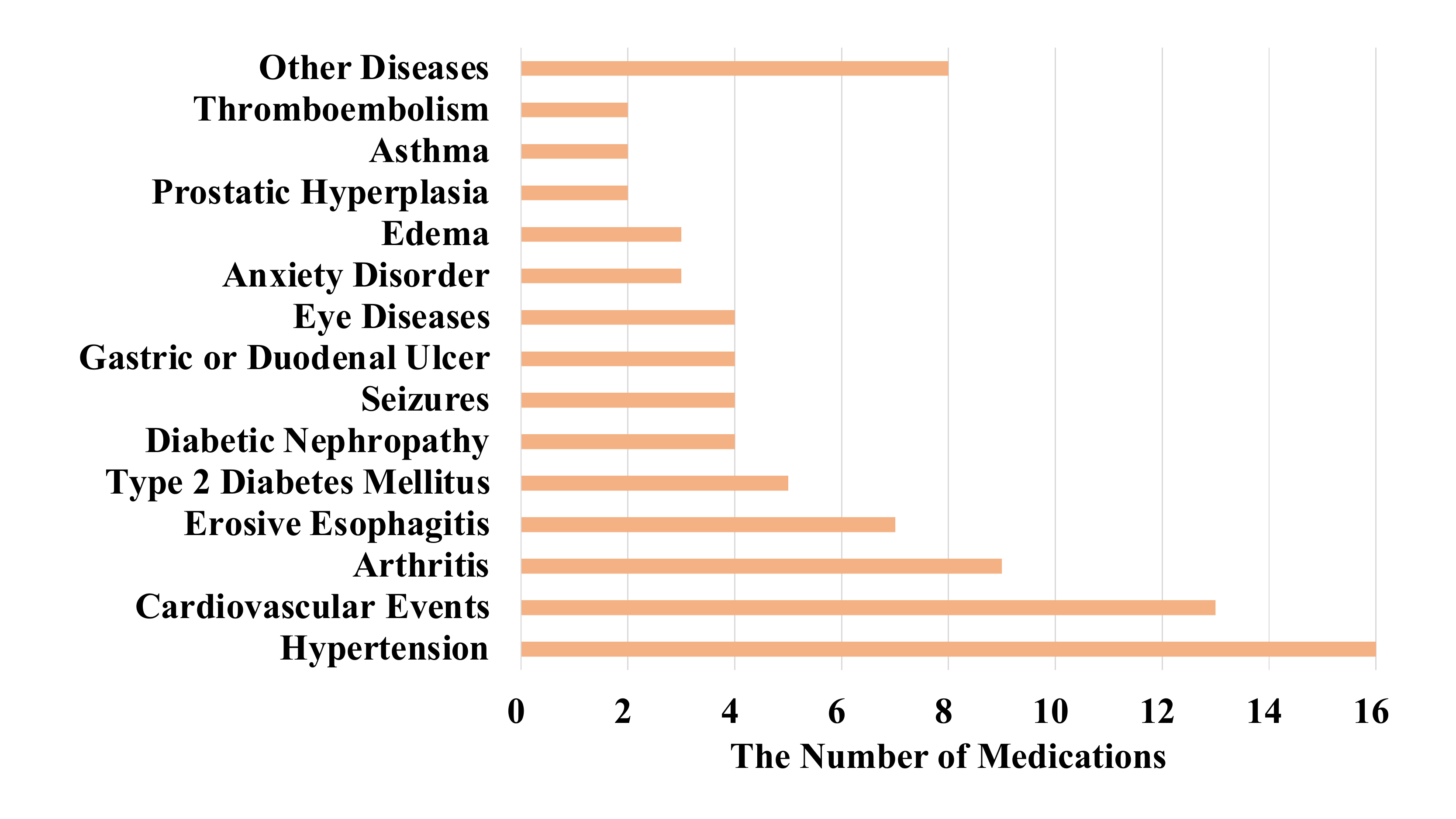}
  \vspace{-0.3cm}
  \caption{The distribution of medications for common chronic diseases.}
  \vspace{-0.5cm}
  \label{drug_dis_Prec}
\end{figure}

%To evaluate the effectiveness of our decision support system, the medications that the participants were taking were collected. 
In total, the participants took 86 medications that are commonly used to treat chronic conditions. For example, Doxazosin, a medication commonly used by the participants, is an alpha-1 adrenergic receptor used to treat mild to moderate hypertension and urinary obstruction due to benign prostatic hyperplasia. Fig. \ref{drug_dis_Prec} shows the distribution of these 86 drugs for various diseases. As there is usually more than one drug available for treating a chronic disease such as diabetes, gastrointestinal diseases, and arthritis, the choice of the most appropriate drug is a significant consideration for doctors, and our proposed DSSDDI is designed to help doctors make decisions more efficiently.
% The detailed names and indications for these 86 medications are listed in Table \ref{tab:druginfo}. 
% Table~\ref{tab:meduse} shows the distribution of the number of medications taken by participants. From Table~\ref{tab:meduse} we can see that nearly half of the total 4157 participants were taking two or more medications. This motivates the design of the DSSDDI, which uses drug-drug interaction to improve the effectiveness of medication suggestion while providing an explanation for the medication suggestions given. 
We collect pre-trained embedding of each drug in the Drug Repurposing Knowledge Graph (DRKG) \cite{drkg2020} as the original feature of the drug for the medication suggestion prediction.
Each pre-trained embedding is trained using a classical knowledge representation learning method named TransE \cite{bordes2013translating} with a dimension size of 400.

% \begin{table}[t]
%   \centering
%   \caption{Distribution of the number of medications taken by participants.}
%     \begin{tabular}{p{9.165em}rrrrrrr}
%     \hline
%     \#Medication Use & 1     & 2     & 3     & 4     & 5     & 6     & 7 \\
%     \hline
%     \#Participants & 2265  & 1192  & 477   & 176   & 41    & 5     & 1 \\
%     \hline
%     \end{tabular}%
%   \label{tab:meduse}%
% \end{table}%

\vspace{-0.2cm}
\subsection{Drug-Drug Interactions}\label{DDIIntro}
DrugCombDB~\cite{liu2020drugcombdb} is a database containing drug-drug interactions obtained from various sources, including external databases, manual curations from PubMed literature and experimental results. We take the drug-drug interactions that have been classified as exhibiting synergistic or antagonistic effects from DrugCombDB\footnote{http://drugcombdb.denglab.org/download/}. For the 86 drugs used for suggestion, we extract 97 drug pairs classified as having synergistic effects and 243 drug pairs classified as having antagonistic effects from the DrugCombDB database. Based on these drug-drug interactions, the proposed decision support system obtains better effectiveness by avoiding pairs with antagonistic effects and promoting pairs with synergistic effects in medication suggestions.

\begin{figure*}[t]
  \centering
  \includegraphics[width=0.98\linewidth]{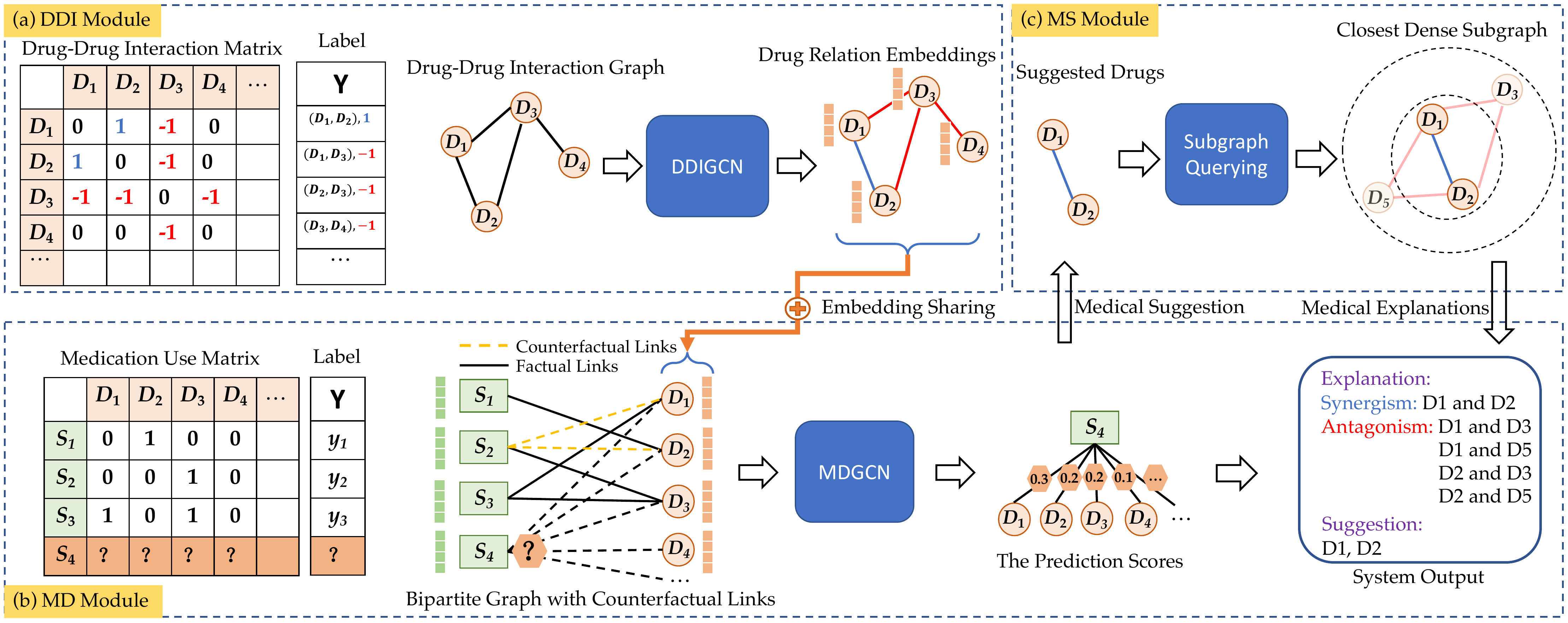}
  \vspace{-0.3cm}
\caption{The proposed decision support system (DSSDDI) consists of three modules: DDI module, MD module and MS module. (a) In the DDI module, we construct the drug-drug interaction matrix where the entries of the matrix represent the synergistic effects (blue) or antagonistic effects (red) among drugs. Through a proposed DDIGCN, we obtain drug relation representations shared with the MD module. (b) In the MD module, we first construct a bipartite graph from the medication use (black lines) and counterfactual links (yellow dashed lines). Then, we obtain the suggested drugs through the proposed MDGCN. (c) In the MS module, the explanation of the suggested drugs is extracted through a subgraph querying algorithm.}
\label{framework}
\vspace{-0.2cm}
\end{figure*}

% \vspace{-0.2cm}
\section{Problem Formulation}\label{Profor}
In this paper, we design three modules for the proposed decision support system: Drug-Drug Interaction (DDI) module, Medical Decision (MD) module and Medical Support (MS) module. In this section, we will first introduce the generalized \textit{decision support system}, then define the \textit{DDI graph} constructed in the DDI module, and give the definitions of \textit{medical decision} and \textit{medical support}.

\vspace{0.1cm}
\begin{myDef}[\textbf{Decision Support System}]
Given a set of drug candidates denoted as $V = \{D_{1}, D_{2}, \cdots, D_{|V|}\}$, the decision support system is designed to suggest a list of appropriate drugs from $V$ to a patient $S_i$ based on the patient features $\mathbf{x}_i$. 
% accompanied by an explanation of that medication suggestion, where $\mathbf{x}_i$ is a vector 
%consisting of the patient's information such as gender, age, blood pressure, etc.
\end{myDef}

\vspace{0.1cm}
\begin{myDef}[{\textbf{Drug-Drug Interaction (DDI) Graph}}]\label{DDIgraph}
We define the drug-drug interaction (DDI) graph as $G=(V, E)$, where the node set $V=\{D_{1}, D_{2}, \cdots, D_{|V|}\}$ denotes the drugs and the edge set $E$ denotes the synergistic or antagonistic effects between drugs. An edge $e_{uv}=1$ in $E$ represents a synergistic effect between drugs $D_u$ and $D_v$, and an edge $e_{uv}=-1$ in $E$ represents an antagonistic effect.
%and pharmacological effects $E$, the drug-drug interaction graph is defined as a set of triplets $G = \{(D_v, e_{vu}, D_u)| D_v \in V, e_{vu}\in E, D_u\in V\}$, where each triplet $(D_v, e_{vu}, D_u)$ represents that drug $D_v$ and drug $D_u$ have pharmacological effect $e_{vu}$.
\end{myDef}

\vspace{0.1cm}
\begin{myDef}[\textbf{Medical Decision}]
The target of medical decision is to identify the most effective combination of drugs for the patients from a set of drug candidates.  By representing the relationship between patients and drugs as a bipartite graph, we formulate the medical decision problem based on the results of link prediction and further use the drug-drug interactions as constraints to identify the appropriate drugs.
% \rr{Wait, what is the definition of mecial decision? May need one sentence to explain.}
% It is different from the knowledge graph-based link prediction problem whose outcome is the score of linking the patient to each drug, since the link prediction ignores the antagonistic effects between drugs.
% Medical decision should be based on the results of the link prediction and further use the Drug-Drug Interaction (DDI) as constraints to identify the appropriate drugs through pruning on the DDI graph.
% \rr{We have not defined the DDI graph!}
Specifically, suppose that we are given a set of observed patients' data denoted as $\mathcal{O} = \{\mathbf{x}_i,\mathbf{y}_i\}^m_{i=1}$, $\mathbf{y}_i$ is a vector with $y_{iv} = 1$ if patient $S_i$ is taking drug $D_v$ and $y_{iv} = 0$ otherwise, $m$ is the number of observed patients. The set of unobserved patients' data is denoted as $\mathcal{U} = \{\mathbf{x}_j,\mathbf{y}_j\}^n_{j=m+1}$, where $n$ is the number of all patients. 
Given the patient features ${\mathbf{x}_j}$, the target of medical decision is first to predict the score of each drug and then to suggest the most $k$ reliable drugs $Q = \{D_{q1}, D_{q2}, \cdots, D_{qk}\}$ to the unobserved patient $S_j$ based on drug-drug interactions.
\end{myDef}

\vspace{0.1cm}
\begin{myDef}[\textbf{Medical Support}]
The target of medical support is to find explainable factors through drug-drug interactions for the $k$ suggested drugs $Q = \{D_{q1}, D_{q2}, \cdots, D_{qk}\}$. Specifically, given $Q$ and the DDI graph $G$, we can find a subgraph $G_{sub}$ of $G$ containing all drug-drug interactions associated with the suggested drugs, and thus can act as medical support for the Medical Decision module.
\end{myDef}

\section{The Proposed DSSDDI}\label{ProDSSDDI}
Fig. \ref{framework} depicts the overall architecture of DSSDDI which consists of three modules: Drug-Drug Interaction, Medical Decision and Medical Support. 
In Drug-Drug Interaction, we learn the drug relation representations. In Medical Decision, we capture the causal relationship between DDI and medication use.
In Medical Support, we generate the explanation of the suggested drugs.
In this section, we elaborate on each module. 

\subsection{Drug-Drug Interaction Module}
In the Drug-Drug Interaction (DDI) module, we first develop a model, DDIGCN, to learn the drug representations. The main idea is to learn drug relation features through synergistic or antagonistic effects between drugs. In the following, we first describe how to construct the DDI graph, then illustrate how to update the drug representations by DDIGCN, and finally describe the model training process.

\vspace{0.1cm}
\subsubsection{DDI Graph Construction}\label{consDDIgraph}
% As stated in Section \ref{DDIIntro}, we collect 97 drug pairs with synergistic effects and 243 drug pairs with antagonistic effects for the 86 drugs used for recommendation. 
As described in Definition \ref{DDIgraph}, we construct the DDI graph $G = (V, E)$ based on the data collected from DrugCombDB~\cite{liu2020drugcombdb} with Drug ID embedding vectors $\mathbf{z}_v$ for $D_v\in V$. To better capture the relation features among drugs, we use one-hot ID embeddings instead of pre-trained embeddings as the original features in this DDI module. \btrevise{Besides synergistic and antagonistic effects, we add the third type of edges between drugs in the DDI graph to explicitly indicate that they have no interactions. Specifically, we randomly sample drug pairs, denoted as $D_u$ and $D_v$, from $V$ with no synergistic or antagonistic effect, and create an edge $e_{uv}=0$ to represent the lack of interactions between them. In this manner, DDIGCN can capture synergistic and antagonistic drug-drug interactions as well as no interactions in drug embeddings.}

%\btrevise{Besides synergistic and antagonistic effects, there are unknown interactions (e.g., interactions that may exist but are not yet discovered, or no interactions between drugs) in reality~\cite{cami2013pharmacointeraction}. To reflect this, we randomly sample drug pairs, denoted as $D_u$ and $D_v$, from $V$ with no synergistic or antagonistic effect between them, and we create an edge $e_{uv}=0$ to represent such unknown interactions. In this manner, DDIGCN can capture synergistic and antagonistic drug-drug interactions as well as unknown interactions in drug embeddings.}

%In addition, \btrevise{besides edges with synergistic effects and edges with antagonistic effects, there are some edges where the DDI relationship is unknown or insignificant, i.e. $e_{vu}=0$. We randomly sampled edges $e_{vu}\notin E$ where $D_v, D_u \in V$ with the same number of edges with synergistic effects plus edges with antagonistic effects to participate in the training as unknown or insignificant edges.}
% we randomly sample $e_{vu}\notin E$ where $D_v, D_u \in V$ and let $e_{vu}=0$. %To keep the data balanced a total of $|V|$ edges are sampled.
\vspace{0.1cm}
\subsubsection{DDIGCN}
In this step, we design a DDIGCN to update the drug representations. We use
Graph Isomorphism Network (GIN)~\cite{xu2018powerful} as the backbone. The graph convolutional operation is defined as:
\vspace{-0.3cm}
\begin{equation}\label{ddirep}
    \mathbf{z}^{(t)}_v = f^{(t)}_{\mathbf{\Theta_1}} \left( (1 + \epsilon^{(t)}) \cdot
        \mathbf{z}^{(t-1)}_v + \frac{\sum_{u\in \mathcal{N}_v} \mathbf{z}^{(t-1)}_u}{|\mathcal{N}_v|} \right),
\vspace{-0.2cm}
\end{equation}
where $\mathbf{z}^{(t)}_v$ denotes the updated hidden representation of drug $D_v$ after $t$ layers propagation, $f^{(t)}_{\mathbf{\Theta_1}}$ denotes the multi-layer perceptrons (MLP) with parameters $\mathbf{\Theta}_1$, $\epsilon^{(t)}$ is a learnable parameter of the $t$-th graph convolutional layer, $\mathcal{N}_v$ denotes the set of drugs that have interactions with drug $D_v$.

\btrevise{Besides GIN, we also consider signed graph-based models, such as SGCN~\cite{DBLP:conf/icdm/Derr0T18}, SiGAT~\cite{DBLP:conf/icann/HuangSHC19} and SNEA~\cite{DBLP:conf/aaai/LiTZC20} as alternative backbones, as there are both positive and negative edges in the DDI graph $G$. Take SGCN as an example, we denote $B_v(t)=\{D_u | e_{uv} = 1\} $ and $U_v(t) = \{D_u | e_{uv} = -1\}$ for drug $D_v$.
\eat{
\begin{equation}
    \begin{split}
        B_v(t) = \{D_u | e_{uv} = 1\} \\
    U_v(t) = \{D_u | e_{uv} = -1\}
    \end{split}
\end{equation}
}
The synergistic and antagonistic hidden representations of drug $D_v$, denoted by $\mathbf{h}_v^{B(t)}$ and $\mathbf{h}_v^{U(t)}$ respectively, are updated as:
\begin{equation}
    \mathbf{h}_v^{B(t)} = \sigma(\mathbf{W}^{B(t)}[\sum\limits_{e_{iv} = 1}\frac{\mathbf{h}_i^{B(t-1)}}{|B(t)|}, \sum\limits_{e_{jv} = -1}\frac{\mathbf{h}_j^{U(t-1)}}{|U(t)|}, \mathbf{h}_v^{B(t-1)}])
\end{equation}
\begin{equation}
    \mathbf{h}_v^{U(t)} = \sigma(\mathbf{W}^{U(t)}[\sum\limits_{e_{iv} = 1}\frac{\mathbf{h}_i^{U(t-1)}}{|U(t)|}, \sum\limits_{e_{jv} = -1}\frac{\mathbf{h}_j^{B(t-1)}}{|B(t)|}, \mathbf{h}_v^{U(t-1)}])
\end{equation}
Finally, $\mathbf{z}^{(t)}_v$ is obtained by concatenating $\mathbf{h}_v^{B(t)}$ and $\mathbf{h}_v^{U(t)}$:
\begin{equation}
   \mathbf{z}^{(t)}_v =  [\mathbf{h}_v^{B(t)}, \mathbf{h}_v^{U(t)}],
\end{equation}
where $\sigma$ is a non-linear activation function, $\mathbf{W}^{B(t)}$ and $\mathbf{W}^{U(t)}$ are the linear transformation matrices responsible for the synergistic and antagonistic interactions, respectively.}

% DDIGCN samples neighboring nodes in the process of updating the drug representation according to the proportion of different types of edges. Since there are more antagonist-containing drug pairs than synergistic ones, we sample more neighboring nodes that constitute antagonists. This is also consistent with the need for a decision support system, as we should avoid drug pairs containing antagonistic effects being suggested to doctors, so more drug pairs containing antagonistic effects should be sampled when updating drug representations.

\subsubsection{Model Training}
{By treating DDIGCN as an edge regression model, we train the model through the Mean Squared Error (MSE) loss function \cite{berger1985certain}.} The score of each edge is calculated by the inner product of two drug representations:
\begin{equation}
    \hat{e}_{vu} = \mathbf{z}_v^{(t)\top}\mathbf{z}^{(t)}_u.
\end{equation}
The MSE loss is defined as:
\begin{equation}
    \mathcal{L}_{M} = \frac{1}{|E_{train}|}\sum_{v,u\in E_{train}} (\hat{e}_{vu}-{e}_{vu})^2,
\end{equation}
where $E_{train}$ denotes all edges involved in training.
% In DSSDDI, we first train DDIGCN to obtain the drug relation representations $\mathbf{z}_v^{(t)}$, then by \textit{Embedding Sharing} we add the drug relation representations $\mathbf{z}_v^{(t)}$ to the drug representations $\mathbf{h}'_v$ in Eq. \eqref{updrug2} proposed below to train the MDGCN.

\subsection{Medical Decision Module}
% In Medication Advice (MA) module, we first update the feature representations of patients and drugs by fully connected layers (FC). Then we construct a bipartite graph based on the observed patients' medication use and utilize Medication Advice Graph Convolutional Network (MDGCN) to further update the drug representations. Next, we calculate the dot product of the patients' hidden representations and the hidden representation of each drug to predict the scores of the patients taking each drug. Finally, an edge pruning operation is employ to  remove the nodes causing antagonistic effects according to the DDI graph. 
In the Medical Decision (MD) module, we construct a \textit{bipartite graph} based on the observed patients’ medication use and develop a Medical Decision Graph Convolutional Network (MDGCN) to provide medical suggestion. To capture the causal relationship between DDI and patient medication use, we augment the graph data with a causal model to generate counterfactual links for training MDGCN.
Correspondingly, this MD module can be described in three parts: Counterfactual Links for Medical Decision, MDGCN and Model Training.

% \subsubsection{Graph Learning with Causal Model}
\vspace{0.1cm}
\subsubsection{Counterfactual Links for Medical Decision}
The target of counterfactual causal inference methods is to capture the causal relationship between treatment and outcome by exploring the counterfactual questions like ``would the outcome be different if the treatment was different?'' \cite{morgan2015counterfactuals}. Thus, given the context, treatments, and corresponding outcomes, we can infer the causal relationship by finding the effect of different treatments on the outcomes.

\begin{figure}[t]
% \vspace{-0.2cm}
\centering
\subfigure[Link prediction with causal model. The treatment $\mathbf{T}_{vu}$ is only related to the representations of node $v$ and node $u$.]{
\label{LP_causal}
% \centering
% \includegraphics[width=0.45\columnwidth]{casual_a.pdf}
\includegraphics[width=0.4\columnwidth]{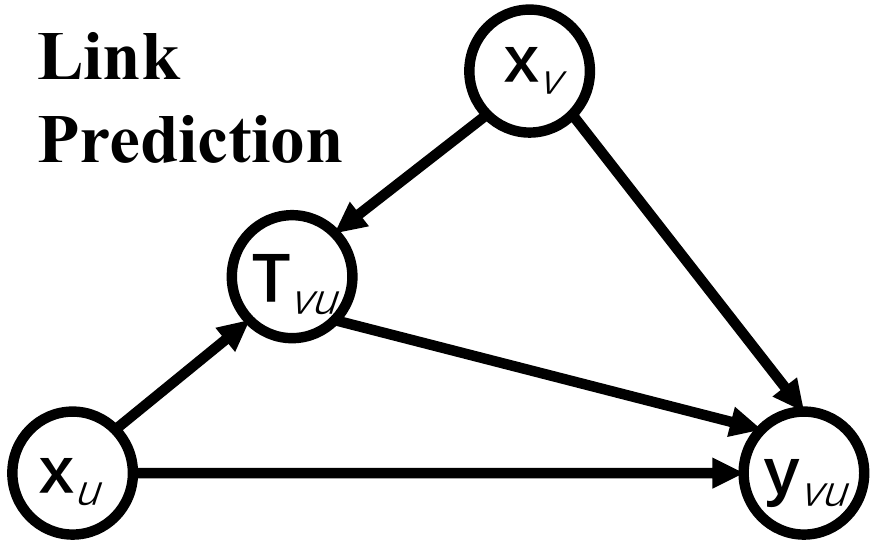}
}
\subfigure[Medical decision with causal model. The treatment $\mathbf{T}_{iv}$ is related to the representations of all the patients $\mathbf{X}$ and drugs $\mathbf{Z}$.]{
\label{MD_causal}
% \centering
% \includegraphics[width=0.45\columnwidth]{casual_b.pdf}
\includegraphics[width=0.4\columnwidth]{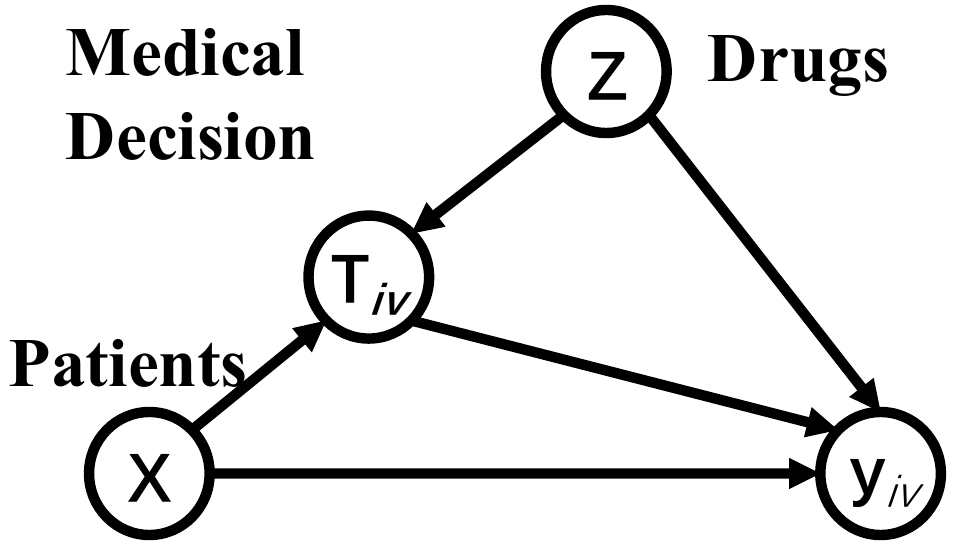}
}
\vspace{-0.1cm}
% \scriptsize
\caption{Causal model improves patient and drug representation learning.}
\vspace{-0.3cm}
\label{causal}
\end{figure}

% Zhao \textit{et al.} \cite{zhao2022learning} leverage counterfactual causal model to improve link prediction by learning form both factual and counterfactual outcomes as observed and augmented data. 
We describe the idea of link prediction with causal model which is an analogy of making medical decision with causal model. Fig. \ref{LP_causal} illustrates link prediction with the causal model \cite{zhao2022learning}, in which the context $\textbf{x}_v$ and $\textbf{x}_u$ are representations of node $v$ and node $u$, the treatment $\textbf{T}$ is defined based on the graph structure information, the outcome $y_{vu}$ is the link existence between node $v$ and node $u$. 
By learning from both factual outcomes $\textbf{Y}$ and the counterfactual outcomes $\textbf{Y}^{CF}$ obtained from the factual treatment $\textbf{T}$ and counterfactual treatment $\textbf{T}^{CF}$, the causal relationship between graph structure and link existence is captured to improve node representations. 
% When the treatment is different, we define the unobserved matrix of counterfactual links $\textbf{Y}^{CF}$ as counterfactual outcomes, while the observed adjacency matrix is denoted by $\textbf{Y}$ as factual outcomes. 

The target of the proposed medical decision problem is to explore ``\textbf{will patients in the same group take drugs with antagonistic effects?}''.
As shown in Fig. \ref{MD_causal}, we consider the representations of patients $\mathbf{X}$ and drugs $\mathbf{Z}$ as context and medication use $\mathbf{Y}$ as outcome, where $\mathbf{X}$ denotes the matrix formed by the combination of all patient features $\{\mathbf{x}_1, \mathbf{x}_2, \cdots\}^\top$, $\mathbf{Z}$ denotes the matrix formed by the combination of all drug features $\{\mathbf{z}_1, \mathbf{z}_2, \cdots\}^\top$ and $\mathbf{Y}$ denotes the matrix formed by the combination of patients' medication use $\{\mathbf{y}_1, \mathbf{y}_2, \cdots\}^\top$.
% However, different from link prediction problem described above, we cannot simply adopt the graph structure as treatment. 
Since the treatment between patient $S_i$ and drug $D_v$ may be influenced by the medical decision of other similar patients as well as the DDI with other drugs, we define treatment $\mathbf{T}_{iv}$ based on the representations of all the patients $\mathbf{X}$ and drugs $\mathbf{Z}$. In summary, the treatment $\mathbf{T}$ is defined in three steps. 
% Different from the treatment defined according to the graph structure in \cite{zhao2022learning}, we define treatment based on patient similarity combined with DDI knowledge. 
Firstly, we define the treatment matrix $\mathbf{T}$ as $\mathbf{T}_{iv} = 1$ if patient $S_i$ and drug $D_v$ have a link in the observed data, and $\mathbf{T}_{iv} = 0$ otherwise. Then we cluster the patients by a clustering method such as $K$-means \cite{hartigan1979algorithm} denoted as $c : \mathbf{X} \rightarrow N$ that outputs the index of cluster that each patient belongs to, where the number of clusters is determined by the number of chronic diseases in the observed data. We set the treatment $\mathbf{T}_{jv} = 1$ if $c(S_i) = c(S_j)$ and $\mathbf{T}_{iv} = 1$ in the second step. In the final step, we assume that treatment $\mathbf{T}_{iu} = 1$ if $e_{vu} = 1$ and $\mathbf{T}_{iv} = 1$ according to the DDI graph in Section \ref{consDDIgraph}.

Then we find the nearest neighbor with the opposite treatment for each patient-drug pair and use the nearest neighbor’s outcome as a counterfactual link. Formally, $\forall (S_i, D_v) \in S \times V$, its counterfactual link $(S_j, D_u)$ is defined as
\begin{equation}\label{cflink}
\begin{aligned}
    (S_j, D_u)= \mathop{\arg\min}\limits_{S_j \in S, D_u \in V}\{\rm{dis}(\mathbf{x}_i, \mathbf{x}_j)+\rm{dis}(\mathbf{z}_v,\mathbf{z}_u)|\\
    \mathbf{T}_{ju}=1-\mathbf{T}_{iv}, \rm{dis}(\mathbf{x}_i, \mathbf{x}_j)<\gamma_p, \rm{dis}(\mathbf{z}_v,\mathbf{z}_u)<\gamma_d\},
\end{aligned}
\end{equation}
where $\rm{dis}(\cdot, \cdot)$ is specified as Euclidean distance, $\mathbf{x}_i\in\mathbb{R}^{d_1}$ and $\mathbf{x}_j\in\mathbb{R}^{d_1}$ denote the original feature of patient $S_i$ and patient $S_j$, $\mathbf{z}_v\in\mathbb{R}^{d_2}$ and  $\mathbf{z}_u\in\mathbb{R}^{d_2}$ denote the original feature of drug $D_v$ and drug $D_u$, $\gamma_p$ and $\gamma_d$ are hyperparameters that define the maximum distance that two patients and two drugs are considered as similar, respectively. Finally, we define the counterfactual treatment matrix $\mathbf{T}^{CF}$ and counterfactual adjacency matrix $\mathbf{Y}^{CF}$ as
\begin{equation}
\begin{aligned}
\mathbf{T}^{CF}_{iv}, y^{CF}_{iv} = \left\{
    \begin{array}{ll}
    1-\mathbf{T}_{iv}, y_{ju} &,\ \rm{if\ } \exists (S_j, D_u) \in S \times V \\ &\rm{satisfies\ Eq.\ }\eqref{cflink};\\
    \mathbf{T}_{iv}, y_{iv} &,\ \rm{ otherwise}.\\
       \end{array}
    \right.
\end{aligned}
\end{equation}

Different from traditional link prediction that only takes the observed outcomes $\mathbf{Y}$ as the training target, we take counterfactual outcomes $\mathbf{Y}^{CF}$ as the augmented training data for MDGCN training.

\vspace{0.1cm}
\subsubsection{MDGCN}

We divide MDGCN into an encoder and a decoder. The encoder is used to update patient and drug representations, and the decoder is used to predict patients' medication use.

\textbf{MDGCN Encoder}: In the initial step, each patient and drug is associated with an original feature. Then two fully connected layers (FC) are leveraged to map the feature representations of all patients and drugs respectively to the same feature dimension as follows:
\begin{equation}\label{upsubject1}
    \mathbf{h}_i = \sigma(\mathbf{W}_1\mathbf{x}_i + \mathbf{b}_1), i = 1, 2, \cdots, m,
\end{equation}
\begin{equation}\label{updrug1}
    \mathbf{h}_v = \sigma(\mathbf{W}_2\mathbf{z}_v + \mathbf{b}_2), v = 1, 2, \cdots, |V|,
\end{equation}
where $h_i$ and $h_v$ denote the hidden representations of patient $S_i$ and drug $D_v$, respectively; $\sigma(\cdot)$ denotes an element-wise activation function, $W_1 \in \mathbb{R}^{d_3\times d_1}$, $W_2 \in \mathbb{R}^{d_3\times d_2}$ and $b_1 \in \mathbb{R}^{d_3}$, $b_2 \in \mathbb{R}^{d_3}$ are learnable parameters of the fully connected layers.

% \subsubsection{Drug Representations Updating}
After mapping the original features of patients and drugs to the same dimension, we use MDGCN to update the hidden representations of drugs. In MDGCN, the feature transformation and nonlinear activation functions are abandoned, it only adopts the simple weighted sum aggregator to update the node features. The graph convolutional operation is defined as:
\begin{equation}\label{mdgcn1}
    \mathbf{h}_i^{(t)} = \sum_{v\in \mathcal{N}_{i}}\frac{1}{\sqrt{|\mathcal{N}_{i}|}\sqrt{|\mathcal{N}_{v}|}}\mathbf{h}_{v}^{(t-1)},
\end{equation}
\begin{equation}\label{mdgcn2}
    \mathbf{h}_v^{(t)} = \sum_{i\in \mathcal{N}_v}\frac{1}{\sqrt{|\mathcal{N}_v|}\sqrt{|\mathcal{N}_i|}}\mathbf{h}_i^{(t-1)},
\end{equation}
where $\mathbf{h}_i^{(t)}$ and $\mathbf{h}_v^{(t)}$ respectively denote the updated hidden representation of patient $S_i$ and drug $D_v$ after $t$ layers propagation, $\mathbf{h}_i^{(0)} $ is equal to $\mathbf{h}_i$ calculated by Eq. \eqref{upsubject1} and $\mathbf{h}_v^{(0)}$ is equal to $\mathbf{h}_v$ calculated by Eq. \eqref{updrug1}, $\mathcal{N}_i$ denotes the set of drugs that are taken by patient $S_i$, $\mathcal{N}_v$ denotes the set of patients that are taking drug $D_v$. 

After $T'$ layers of graph convolutional operation, the hidden representations obtained at each layer are combined to form the final representation of drug $D_v$:
\vspace{-0.2cm}
\begin{equation}\label{updrug2}
    \mathbf{h}'_v = \sum_{t=0}^{T'}\beta_t\mathbf{h}_v^{(t)},
\vspace{-0.1cm}    
\end{equation}
where $\beta_t \geq 0 $ is a hyperparameter to represent the importance of the $t$-th layer representation in constituting the final representation. \btrevise{Note that the target node's feature will be added at $t=0$, so we don't need to add the target node's feature in Eq.~\eqref{mdgcn1} and~\eqref{mdgcn2}.} Since representations at different layers capture different semantics, e.g., the first layer smooths patients and drugs that have interactions, the second layer smooths drugs that have overlap with interacted patients, and higher-layers capture higher-order proximity~\cite{he2020lightgcn}. Hence, this layer combination operation will make the drugs' representations more similar to those of patients with their corresponding diseases.
% In DSSDDI, we first train DDIGCN to obtain the drug relation representations $\mathbf{z}_v^{(t)}$, then by \textit{Embedding Sharing} we add the drug relation representations $\mathbf{z}_v^{(t)}$ to the drug representations $\mathbf{h}'_v$ in Eq. \eqref{updrug2} proposed below to train the MDGCN.

% \subsubsection{Personalized Medication Use Prediction}
\textbf{MDGCN Decoder}: As shown in Fig. \ref{drug_dis_Prec}, there are many drugs to treat the same type of disease, for example, Doxazosin, Terazosin and Prazosin can all be used to treat hypertension, so how to personalize and suggest more appropriate drugs according to the features of the patient is the challenge to be solved in this paper.
To obtain personalized medication suggestions, we adopt the hidden representation $\mathbf{h}_i$ obtained from Eq. \eqref{upsubject1} of the patient $S_i$ to predict his/her medication use. Compared with the patient representations obtained after MDGCN, the patient representations before MDGCN are more differentiated because there is no over-smoothing of patient representations resulting from aggregation of similar drug representations. In Section \ref{exp_result}, we compare the personalization of patient representations before and after MDGCN experimentally.

% In this paper, the difference between DRDDI and DR is whether to add DDI relation representations obtained in Eq. \eqref{ddirep} in this step, i.e. $\mathbf{h}'_v = \mathbf{h}'_v + \mathbf{z}_v $.
Next, we add DDI relation representations shared in Eq. \eqref{ddirep} to the final drug representation $\mathbf{h}'_v$, i.e., $\mathbf{h}'_v = \mathbf{h}'_v + \mathbf{z}_v $. Based on MLP, the encoder of MDGCN is defined as:
\begin{equation}
    \hat{y}_{iv} = f^{(t)}_{\mathbf{\Theta_2}}([\mathbf{h}_i^\top\odot \mathbf{h}'_v,\mathbf{T}_{iv}]),
\end{equation}
\begin{equation}
    \hat{y}^{CF}_{iv} = f^{(t)}_{\mathbf{\Theta_2}}([\mathbf{h}_i^\top\odot \mathbf{h}'_v,\mathbf{T}^{CF}_{iv}]),
\end{equation}
\vspace{-0.1cm}
where $f^{(t)}_{\mathbf{\Theta_2}}$ denotes the MLP with parameters $\mathbf{\Theta}_2$, $[\cdot, \cdot]$ represents the concatenation of vectors, $\odot$ represents Hadamard Product.

% \btrevise{\textbf{Edge Pruning}}: At last, we adopt an edge pruning strategy to avoid antagonistic effect between the suggested drugs. That is, each drug is sequentially selected into the set of suggested drug candidates $Q$ by prediction scores, and then based on the DDI graph, it is verified whether there is an antagonistic effect between the newly selected drug $D_k$ and other already selected drugs $\{D_1, D_2, \cdots, D_{k-1}\}$, and if so, the newly selected drug is deleted, and if not, it is retained.

\textbf{System Output}: At last, we combine the \textit{medical suggestion} and the corresponding \textit{medical explanations} obtained through the MS module, thus constructing the system output to be displayed to the doctors.

\begin{figure}[t]
  \centering
  \includegraphics[width=\linewidth]{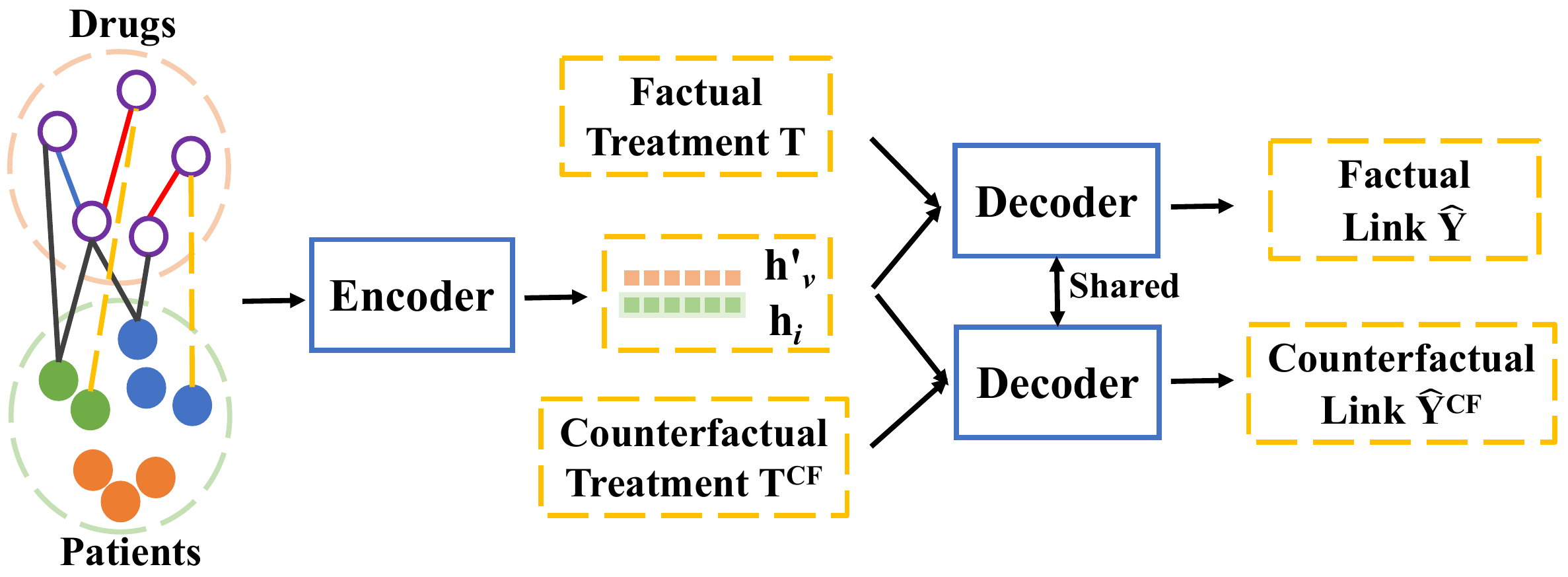}
  \vspace{-0.4cm}
  \caption{Given the corresponding factual and counterfactual treatments, MDGCN is trained to predict factual and counterfactual links. The different colored patients indicate that they belong to different clusters. The red (blue) lines between drugs indicate antagonistic (synergistic) effects. The black~(yellow) lines between patients and drugs indicate factual (counterfactual) links.}
  \vspace{-0.4cm}
  \label{MDGCN}
\end{figure}
\subsubsection{Model Training} 
% The trainable parameters of MD module are only the weight matrices $\textbf{W}_1$, $\textbf{W}_2$ and the biases $\textbf{b}_1$, $\textbf{b}_2$ in the Section \textit{4.2.1}.
The training process of MDGCN is shown in Fig. \ref{MDGCN}. During the model training, we predict factual links and counterfactual links and optimize towards $\mathbf{Y}$ and $\mathbf{Y}^{CF}$, respectively.
We adopt 1:1 negative sampling to sample negative edges for training, and then train the model by the cross-entropy loss functions defined as:
\vspace{-0.1cm}
\begin{equation}
    \mathcal{L}_{C}=-\sum_i^m\sum_{v\in \mathcal{N}_i}[y_{iv}\mathrm{log}(\hat{y}_{iv})+(1-y_{iv})\mathrm{log}(1-\hat{y}_{iv})],
\end{equation}
\vspace{-0.2cm}
\begin{equation}
    \mathcal{L}_{CF}=-\sum_i^m\sum_{v\in \mathcal{N}_i}[y^{CF}_{iv}\mathrm{log}(\hat{y}^{CF}_{iv})+(1-y^{CF}_{iv})\mathrm{log}(1-\hat{y}^{CF}_{iv})].
\vspace{-0.2cm}
\end{equation}
The overall training loss of our model is:
\vspace{-0.1cm}
\begin{equation}
    \mathcal{L}=\mathcal{L}_{C} + \delta \mathcal{L}_{CF},
\vspace{-0.1cm}
\end{equation}
where $\delta$ is a hyperparameter to control the weight of counterfactual outcome estimation loss.

\subsection{Medical Support Module}\label{expDSSDDI}
Given a set of suggested drugs to a patient by the decision support system, we may want to know further why the system suggests these drugs to him/her, or to what extent this suggestion makes sense. To this end, in the Medical Support (MS) module, we utilize a subgraph querying algorithm to find the \textbf{closest dense subgraph} containing the suggested drugs in DDI. With this subgraph, we define a \textbf{Suggestion Satisfaction} measurement, analyze the synergistic and antagonistic interactions between the suggested drugs and explain our suggestions. In this module, we first introduce the definition of the closest dense subgraph, then illustrate the algorithm to find the closest dense subgraph in DDI. At last, we define the Suggestion Satisfaction measurement to explain the suggestions.

\vspace{0.1cm}
\subsubsection{Problem Definitions} 

% We ignore the pos/neg interactions between drugs and consider the DDI as a simple graph $G=(V,E)$. Then in this subgraph, we can further analyze the different interactions between drugs. 
While there are many choices of closeness on graphs, in this work,
we use the closest truss community definition \cite{DBLP:journals/pvldb/HuangLYC15} based on triangles to find the {\em closest dense subgraph} first. 

\vspace{0.1cm}
\begin{myDef}
[\textbf{$p$-truss}]
A triangle in $G$ is a cycle of length 3.  Let $D_u, D_v, D_w \in V$ be the three vertices on the cycle, and we denote this triangle by $\triangle_{uvw}$. Then the support of an edge $e_{uv}$ $\in E$ in $G$, denoted by $sup(e_{uv}, G)$, is defined as the number of triangles in $G$ containing $e_{uv}$, i.e., $|\{\triangle_{uvw}: D_w\in V\}|$. 
% When the context is obvious, we replace $sup(e, G)$ by $sup(e)$. 
A subgraph is $p$-truss if the support number of all the edges is no less than $(p-2)$. The truss number of an edge is defined as the maximum value of $p$ that this edge can contain in the $p$-truss subgraph.
\end{myDef}

Then the closest truss community is formally defined as:
\vspace{0.1cm}
\begin{myDef}
[\textbf{Closest Truss Community (CTC)}]
Given a graph $G$ and a set of query nodes $Q$, $G'$ is a closest truss community (CTC), if $G'$ satisfies the following two conditions: 
(1) \textit{Connected $p$-Truss}: $G'$ is a connected $p$-truss containing $Q$ with the largest $p$, i.e., $Q \subseteq G' \subseteq G$ and $\forall e \in E(G'), sup(e, G') \leq (p -2)$; 
(2) \textit{Smallest Diameter}: $G'$ is a subgraph of smallest diameter satisfying condition (1). That is, $\nexists G'' \subseteq G'$, such that $diam(G'') \le diam(G')$, and $G''$ satisfies condition (1), where $diam(G')$ denotes the diameter of $G'$.

\end{myDef}

In our task, given a set of suggested drugs $Q = \{{D_q}_1, {D_q}_2, \dots, {D_q}_k\}$, we find the closest truss community in the DDI graph which contains all the suggested drugs, and these drugs in the subgraph connect densely through either synergistic or antagonistic interactions. In such a discovered subgraph, we can analyze the drug interactions and explain why we suggest these drugs.

\vspace{0.1cm}
\subsubsection{Subgraph Querying} 

To find out this closest dense subgraph, we use the community search algorithms in \cite{DBLP:journals/pvldb/HuangLYC15}. In particular, it first computes a Steiner tree and then extends the Steiner tree into a dense subgraph. In this subgraph, we find a closest dense subgraph. The main algorithms used in this process include \textit{Steiner Tree Computation} and \textit{Truss Decomposition}. We briefly describe them below.

\vspace{0.1cm}
\paragraph{Steiner Tree Computation} Given a graph $G$ and query nodes $Q$, it firstly constructs a complete distance graph $G'$ of query nodes where the distance is the truss distance defined in \cite{DBLP:journals/pvldb/HuangLYC15}. It then finds a minimum spanning tree $T_s$ of $G'$ , and then constructs another graph $H$ by replacing each edge of tree $T_s$ by its corresponding shortest path in $G'$, and finally finds a minimum spanning tree of $H$ and deletes leaf edges. The detailed algorithm can be found in \cite{mehlhorn1988faster}.

\vspace{0.1cm}
\paragraph{Truss Decomposition} To compute the truss number of each edge, we use the truss decomposition algorithm \cite{WangC12}. It first computes the support number for each edge and sorts all the edges in ascending order of their support. Then it deletes the edges with the smallest support and updates other edges' support. The truss number of one edge is the updated support number when it is deleted. After all the edges are deleted, the truss number of each edge is computed.

The entire algorithm is listed in Algorithm~\ref{algo:explain}. We first do truss decomposition \cite{WangC12} on the DDI graph $G$ (line 1), then use the truss distance function \cite{DBLP:journals/pvldb/HuangLYC15} to compute the Steiner Tree $T_s$ \cite{mehlhorn1988faster} (line 2). Based on $T_s$, we extend it to a dense subgraph $G_0'$ that the truss number of each edge is no less than the minimum truss number of $T_s$ (line 3-7). On this dense subgraph $G_0'$, we do truss decomposition (line 8) and find the connected $p$-truss community with maximum $p$ it can find (line 9). Then we iteratively shrink the subgraph by deleting the furthest nodes while maintaining the truss property (line 10-14). The final closest related subgraph is the subgraph with the smallest diameter during the iterations (line 15).

\begin{algorithm}[t]
\footnotesize
\caption{Subgraph Querying} \label{algo:explain}
\begin{flushleft}
\footnotesize
\textbf{Input:} DDI graph: $G=(V, E)$, \\
\hspace{0.75cm} suggested drugs: $Q =\{{D_q}_1,{D_q}_2,\dots, {D_q}_k\}$ \\
\textbf{Output:} related subgraph of suggested drugs: $G_{sub}=\{{D_1},{D_2},\dots\}$.\\
\end{flushleft}
% \vspace{0.1cm}
\begin{algorithmic}[1]
\footnotesize

\STATE Do truss decomposition on $G$

\STATE  Compute the Steiner Tree $T_s$ containing suggested drugs

\STATE $G_0'\leftarrow T_s$

% \STATE  Expand the Steiner Tree T to a graph $G'$ by involving local related drugs of suggested drugs $D$.
\STATE $p' \leftarrow \min_{e' \in T_s}truss(e')$:

\STATE while size of $G_0'<n_0$

\STATE  \hspace{0.3cm} \textbf{if} ($e$ is an adjacent edge to $G_0'$ and $truss(e)\geq p' $)

\STATE   \hspace{0.3cm} \hspace{0.3cm}  $G_0'\leftarrow G_0' \cup e$

\STATE  Do truss decomposition on $G_0'$ 

\STATE  Find maximum connected $p$-truss subgraph containing $Q$ in $G_0'$

\STATE  $i\leftarrow 0$

% \STATE  Remove the furthest drugs from suggested drugs iteratively and find the most related sugb graph $G_sub$ 
\STATE  while (connected($Q$))

\STATE  \hspace{0.3cm} $G'_{i+1}\leftarrow$ delete the furthest nodes of $G'_i$
\STATE  \hspace{0.3cm} Maintain $p$-truss property of $G'_{i+1}$ 
\STATE  \hspace{0.3cm} $i\leftarrow i+1$

\STATE  \textbf{return} $G_{sub}\leftarrow \arg\min_{G' \in \{G'_{0},G'_{1}, ...\}} dist(G',Q)$

\end{algorithmic}
\end{algorithm}

\vspace{0.1cm}
\subsubsection{Measurement} 
For a suggestion with $k$ drugs, we use Suggestion Satisfaction (\textit{SS}) to measure its rationality.
\vspace{0.1cm}
\begin{myDef}
[\textbf{Suggestion Satisfaction (SS)}]
\eat{\textit{SS} is defined as the sum of the normalized ratio of synergistic edges to the antagonistic edges between the suggested drugs and the normalized ratio of antagonistic edges between the suggested and non-suggested drugs.} Let the closest dense subgraph of the DDI graph w.r.t. the suggested drugs be $G_{sub}$ with $n'$ nodes. $r^{in}_{pos}$ and $r^{in}_{neg}$ denote the number of synergistic edges and antagonistic edges between the $k$ suggested drugs respectively. $r^{out}_{neg}$ is the number of antagonistic edges between the $k$ suggested drugs and $n' - k$ non-suggested drugs. \textit{SS} is defined as follows:
\begin{equation}\label{SS}
    SS=\alpha\frac{2(r^{in}_{pos}+1)}{(r^{in}_{neg}+1)(k(k-1)+2)} + (1-\alpha)\frac{r^{out}_{neg}}{k(n' - k)},
\end{equation}
where the first term explains the synergistic effect between the suggested drugs, while the second explains the antagonistic effect between the suggested drugs and the non-suggested drugs. $\alpha\in (0,1)$ is a hyperparameter that balances the two terms. We expect better synergy between the suggested drugs and greater antagonism with the non-suggested drugs, so a larger \textit{SS} can denote a more appropriate drug suggestion. 
\end{myDef}

The MS module explains the medication suggestions from a visual perspective (subgraphs) and a numerical perspective (SS), giving the doctors a more reliable, easier to understand and more visual explanation of the suggestion.

%Version 2
% It is defined as a normalized ratio of synergistic edges divided by the antagonistic edges between the suggested drugs. Specifically, assume that the smallest subgraph of the DDI graph containing top-$k$ suggested drugs is $G_{sub}$, where there are $r^{in}_{pos}$ synergistic edges and $r^{in}_{neg}$ antagonistic edges between $k$ suggested drugs. Suggestion satisfaction is defined as follows:
% \begin{equation}
%     SS=\frac{2(r^{in}_{pos}+1)}{(r^{in}_{neg}+1)(k(k-1)+2)}.
% \end{equation}
% \end{myDef}

% \vspace{-0.55cm}
\section{Experiments}\label{exp}
In our experiments, we first evaluate the performance of DSSDDI for personalized drug suggestions and compare it with baseline methods. Then we conduct ablation study to prove the superiority of DDIGCN. Next, we demonstrate the explainablility of DSSDDI with the measurement of \textit{SS}. \btrevise{Finally, we validate the model's effectiveness on a public diagnostic data set MIMIC-III~\cite{johnson2016mimic}}.

\vspace{-0.15cm}
\subsection{Experimental Setup}
\vspace{-0.1cm}
\subsubsection{Baselines} We compare DSSDDI with the following baselines.

% \rr{Maybe we can use itemize to make this part more clear? [leftmargin=*] can greatly save the space.}
\textbf{Traditional Methods:}
\begin{itemize}
\item \textbf{UserSim}: Given the problem formulation of our decision support system, we evaluate the effectiveness of the system by predicting the medication use of the unobserved patients, who are not involved in the model training process. Hence, we design a naive baseline method, called User Similarity (\textbf{UserSim}). The suggestion scores of each drug for unobserved patients are obtained by weighting the medication use of each observed patient using the similarity between each observed patient and unobserved patients as weights, and calculated by the following equation:
\vspace{-0.2cm}
\begin{equation}
    \mathbf{Y}_\mathcal{U}=\text{cosine\_similarity}(\mathbf{X}_\mathcal{U},\mathbf{X}_\mathcal{O})\cdot \mathbf{Y}_\mathcal{O},
\vspace{-0.2cm}
\end{equation}
where $\mathbf{Y}_\mathcal{U}$ denotes the prediction score matrix for unobserved patients, $\mathbf{X}_\mathcal{O}$ and $\mathbf{X}_\mathcal{U}$ denote the feature matrix of observed patients and unobserved patients, and $\mathbf{Y}_\mathcal{O}$ denotes the medication use of observed patients.
\item \textbf{ECC} \cite{read2009classifier}: Ensemble Classifier Chain (ECC) is a popular multi-label classification method that models the correlation between labels by feeding both input and prediction by the previous classifier into the next classifier. We employ Logistic Regression (LR) \cite{wright1995logistic} as binary classifiers for each label.
\item \textbf{SVM} \cite{suykens1999least}: Bao \textit{et al.} \cite{bao2016intelligent} demonstrate the good performance of support vector machine (SVM) as a traditional machine learning method for medication suggestion.
\end{itemize}

\textbf{Graph Learning-based Methods:}
\begin{itemize}
\item \textbf{GCMC} \cite{berg2017graph}: A recommendation system organizes historical user behaviors as a holistic interaction graph and employs a GCN encoder to generate representations. GCMC focuses on explicit feedback, constructs multiple adjacency matrices according to
the type of score, and uses different weight matrices to decode different types of edges. 
%In this paper, we use implicit feedback to simplify it to a single-layer GCN model with a bilinear decoder. 
\item \textbf{LightGCN} \cite{he2020lightgcn}: This method simplifies the embedding propagation process by eliminating the nonlinear activation function and feature transformation matrix to obtain a light GCN model.
\btrevise{
\item \textbf{SafeDrug} \cite{DBLP:conf/ijcai/YangXMGS21}: A model equipped with a global Message Passing Neural Network (MPNN) module to capture drugs’ molecule structures and a local bipartite learning module to explicitly model drug-drug interactions. It employs Gated Recurrent Unit (GRU)~\cite{chung2014empirical} to encode patients' features from patients' past visits.
}
\btrevise{
\item \textbf{Bipar-GCN} \cite{DBLP:conf/icde/JinZ00W20}: This method is specifically designed for bipartite graph. The patient embeddings and drug embeddings are obtained by training two structurally identical neural networks called patient-oriented NN and drug-oriented NN with different parameters, respectively.
\item \textbf{CauseRec}~\cite{DBLP:conf/sigir/ZhangYZC021}: A causal recommendation model that learns patient representations by generating counterfactual patient behavior sequences from patients' past visits.}
\end{itemize}
\btrevise{\textbf{Variants of DSSDDI:}
\begin{itemize}
    \item \textbf{DSSDDI(GIN)}: DSSDDI with a common GCN model GIN~\cite{xu2018powerful} as backbone.
    \item \textbf{DSSDDI(SGCN)}: DSSDDI with a signed GCN model SGCN~\cite{DBLP:conf/icdm/Derr0T18} as backbone.
    \item \textbf{DSSDDI(SiGAT)}: DSSDDI with an attention-based signed GCN model SiGAT~\cite{DBLP:conf/icann/HuangSHC19} as backbone.
    \item \textbf{DSSDDI(SNEA)}: DSSDDI with an attention-based signed GCN model SNEA~\cite{DBLP:conf/aaai/LiTZC20} as backbone.
\end{itemize}
}

% To solve the user cold-start problem, we add a fully-connected layer before the graph convolution layer of the GCMC and LightGCN models as Eq. \eqref{upsubject1}, and predict the new subjects’ medication use in the validation set and test set based on the features obtained from the fully-connected layer.

% (6) \textbf{DR}: In addition, we also make a comparison with the DR model that does not incorporate the DDI relation representations.

\vspace{0.1cm}
\subsubsection{Metrics}
We split all patients into training set, validation set and test set in the ratio of 5:3:2, the hyperparameter selection is based on the prediction performance on the validation set. We adopt SS@$k$ (top-$k$ Suggestion Satisfaction) defined in Eq.~\eqref{SS} to assess the explainability of drug suggestions for each method. We also use \btrevise{Precision@$k$}, Recall@$k$ and NDCG@$k$ (Normalized Discounted Cumulative Gain) \cite{jarvelin2002cumulated} to measure the effectiveness of all methods:

% \begin{itemize}

(1) \btrevise{Precision@$k$} and Recall@$k$ are defined as:
\vspace{-0.1cm}
\btrevise{
\begin{equation}
Precision@k=\frac{\sum_j|P(j)\cap Q(j)|}{\sum_j|P(j)|},
\end{equation}
}
\vspace{-0.1cm}
\begin{equation}
Recall@k=\frac{\sum_j|P(j)\cap Q(j)|}{\sum_j|Q(j)|},
\end{equation}
\vspace{-0.1cm}
where $P(j)$ represents the set of the $k$ drugs suggested to the patient $S_j$, $Q(j)$ represents the set of drugs that the patient $S_j$ is taking.

(2) NDCG@$k$ is defined as:
\vspace{-0.1cm}
\begin{equation}
    NDCG@k =\frac{1}{|\mathcal{U}|} \sum_{j\in\mathcal{U}} \frac{DCG_j@k}{IDCG_j},
\end{equation}
\vspace{-0.1cm}
where $DCG_j@k$ is the Discounted Cumulative Gain of the $k$ drugs suggested to the patient $S_j$, and $IDCG_j$ is the ideal DCG for the patient $S_j$. $DCG_j@k$ is defined as:
\vspace{-0.1cm}
\begin{equation}
    DCG_{j}@k=\sum_{s=1}^k \frac{2^{rel_s}-1}{\mathrm{log}_2(s+1)},
\vspace{-0.1cm}
\end{equation}
where $rel_s$ is the graded relevance of the result at position $s$. NDCG@$k$ takes a value between 0 and 1, and the larger the value the greater the suggestion scores for those drugs the patient is taking.
% \end{itemize}

% Table generated by Excel2LaTeX from sheet 'final_resp'
\vspace{-0.1cm}
\begin{table*}[htbp]
  \centering
  \caption{Medication suggestion performance comparison between the proposed method and baseline methods on Chronic data set (the best results are in bold and the second results are underlined).}
    \begin{tabular}{llllllllll}
    \toprule
    Method & \textcolor{black}{Precision@6} & Recall@6 & NDCG@6 & \textcolor{black}{Precision@5} & Recall@5 & NDCG@5 & \textcolor{black}{Precision@4} & Recall@4 & NDCG@4 \\
    \midrule
    UserSim & \textcolor{black}{0.0982 } & 0.2227  & 0.1432  & \textcolor{black}{0.0971 } & 0.2209  & 0.1426  & \textcolor{black}{0.0977 } & 0.2181  & 0.1418  \\
    ECC   & \textcolor{black}{0.0214 } & 0.0537  & 0.0328  & \textcolor{black}{0.0252 } & 0.0519  & 0.0321  & \textcolor{black}{0.0060 } & 0.0108  & 0.0127  \\
    SVM   & \textcolor{black}{0.0670 } & 0.2166  & 0.2062  & \textcolor{black}{0.0635 } & 0.1681  & 0.1847  & \textcolor{black}{0.0787 } & 0.1653  & 0.1838  \\
    \midrule
    GCMC  & \textcolor{black}{0.1362 } & 0.5310  & 0.3652  & \textcolor{black}{0.1447 } & 0.4541  & 0.3181  & \textcolor{black}{0.1638 } & 0.4057  & 0.3146  \\
    LightGCN & \textcolor{black}{0.2073 } & 0.7348  & 0.6012  & \textcolor{black}{0.2358 } & 0.7245  & 0.5681  & \textcolor{black}{0.2581 } & 0.6509  & 0.5436  \\
    \textcolor{black}{SafeDrug} & \textcolor{black}{0.0863 } & \textcolor{black}{0.3098 } & \textcolor{black}{0.2267 } & \textcolor{black}{0.1000 } & \textcolor{black}{0.2952 } & \textcolor{black}{0.2227} & \textcolor{black}{0.1250 } & \textcolor{black}{0.2952 } & \textcolor{black}{0.2233 } \\
    \textcolor{black}{Bipar-GCN} & \textcolor{black}{0.1741 } & \textcolor{black}{0.6267 } & \textcolor{black}{0.4817 } & \textcolor{black}{0.1952 } & \textcolor{black}{0.5911 } & \textcolor{black}{0.4667 } & \textcolor{black}{0.2172 } & \textcolor{black}{0.5363 } & \textcolor{black}{0.4418 } \\
    \textcolor{black}{CauseRec} & \textcolor{black}{0.1707 } & \textcolor{black}{0.1025 } & \textcolor{black}{0.5117 } & \textcolor{black}{0.1124 } & \textcolor{black}{0.4492 } & \textcolor{black}{0.3030 } & \textcolor{black}{\textbf{0.3186 }} & \textcolor{black}{0.2468 } & \textcolor{black}{0.1799 } \\
    \midrule
    \textcolor{black}{DSSDDI(SiGAT)} & \textcolor{black}{0.2214 } & \textcolor{black}{0.8215 } & \textcolor{black}{0.6482 } & \textcolor{black}{0.2514 } & \textcolor{black}{0.7834 } & \textcolor{black}{0.6323 } & \textcolor{black}{0.2876 } & \textcolor{black}{0.7266 } & \textcolor{black}{0.6076 } \\
    \textcolor{black}{DSSDDI(SNEA)} & \textcolor{black}{0.2192 } & \textcolor{black}{0.7854 } & \textcolor{black}{0.5949 } & \textcolor{black}{0.2447 } & \textcolor{black}{0.7364 } & \textcolor{black}{0.5744 } & \textcolor{black}{0.2740 } & \textcolor{black}{0.6684 } & \textcolor{black}{0.5442 } \\
    DSSDDI(GIN) & \textcolor{black}{\underline{0.2272} } & \underline{0.8407}  & \underline{0.6836}  & \textcolor{black}{\underline{0.2534} } & \underline{0.8104}  & \textbf{0.6873 } & \textcolor{black}{0.2900 } & \underline{0.7704}  & \underline{0.6575}  \\
    \textcolor{black}{DSSDDI(SGCN)} & \textcolor{black}{\textbf{0.2348 }} & \textcolor{black}{\textbf{0.8521 }} & \textcolor{black}{\textbf{0.6850 }} & \textcolor{black}{\textbf{0.2670 }} & \textcolor{black}{\textbf{0.8153 }} & \textcolor{black}{\underline{0.6717} } & \textcolor{black}{\underline{0.3077} } & \textcolor{black}{\textbf{0.7746 }} & \textcolor{black}{\textbf{0.6680 }} \\
    \midrule
    \midrule
    Method & \textcolor{black}{Precision@3} & Recall@3 & NDCG@3 & \textcolor{black}{Precision@2} & Recall@2 & NDCG@2 & \textcolor{black}{Precision@1} & Recall@1 & NDCG@1 \\
    \midrule
    UserSim & \textcolor{black}{0.0970 } & 0.1970  & 0.1324  & \textcolor{black}{0.1370 } & 0.1348  & 0.1033  & \textcolor{black}{0.0889 } & 0.0088  & 0.0108  \\
    ECC   & \textcolor{black}{0.0072 } & 0.0098  & 0.0123  & \textcolor{black}{0.0108 } & 0.0098  & 0.0135  & \textcolor{black}{0.0192 } & 0.0094  & 0.0204  \\
    SVM   & \textcolor{black}{0.1050 } & 0.1649  & 0.1863  & \textcolor{black}{0.1575 } & 0.1639  & 0.1970  & \textcolor{black}{0.3029 } & 0.1552  & 0.2536  \\
    \midrule
    GCMC  & \textcolor{black}{0.1791 } & 0.3437  & 0.2898  & \textcolor{black}{0.1971 } & 0.2392  & 0.2406  & \textcolor{black}{0.1815 } & 0.1428  & 0.2344  \\
    LightGCN & \textcolor{black}{0.2925 } & 0.5854  & 0.5436  & \textcolor{black}{0.3347 } & 0.4021  & 0.4187  & \textcolor{black}{0.4231 } & 0.2605  & 0.3786  \\
    \textcolor{black}{SafeDrug} & \textcolor{black}{0.1206 } & \textcolor{black}{0.2182 } & \textcolor{black}{0.1872 } & \textcolor{black}{0.1280 } & \textcolor{black}{0.1536 } & \textcolor{black}{0.1609 } & \textcolor{black}{0.1406 } & \textcolor{black}{0.0902 } & \textcolor{black}{0.1406 } \\
    \textcolor{black}{Bipar-GCN} & \textcolor{black}{0.2484 } & \textcolor{black}{0.4671 } & \textcolor{black}{0.4118 } & \textcolor{black}{0.2861 } & \textcolor{black}{0.3672 } & \textcolor{black}{0.3734 } & \textcolor{black}{0.3377 } & \textcolor{black}{0.2197 } & \textcolor{black}{0.3377 } \\
    \textcolor{black}{CauseRec} & \textcolor{black}{0.2122 } & \textcolor{black}{0.1595 } & \textcolor{black}{0.1064 } & \textcolor{black}{0.1665 } & \textcolor{black}{0.1250 } & \textcolor{black}{0.2494 } & \textcolor{black}{0.1484 } & \textcolor{black}{0.1484 } & \textcolor{black}{0.1484 } \\
    \midrule
    \textcolor{black}{DSSDDI(SiGAT)} & \textcolor{black}{0.3361 } & \textcolor{black}{0.6519 } & \textcolor{black}{0.5745 } & \textcolor{black}{0.3912 } & \textcolor{black}{0.5261 } & \textcolor{black}{0.5214 } & \textcolor{black}{0.4531 } & \textcolor{black}{0.3206 } & \textcolor{black}{0.4531 } \\
    \textcolor{black}{DSSDDI(SNEA)} & \textcolor{black}{0.3133 } & \textcolor{black}{0.5795 } & \textcolor{black}{0.5059 } & \textcolor{black}{0.3365 } & \textcolor{black}{0.4242 } & \textcolor{black}{0.4375 } & \textcolor{black}{0.4038 } & \textcolor{black}{0.2667 } & \textcolor{black}{0.4038 } \\
    DSSDDI(GIN) & \textcolor{black}{\underline{0.3554} } & \underline{0.6918}  & \underline{0.6256}  & \textcolor{black}{\underline{0.4261} } & \textbf{0.5926 } & \underline{0.5842}  & \textcolor{black}{\underline{0.4916} } & \textbf{0.3989 } & \textbf{0.5565 } \\
    \textcolor{black}{DSSDDI(SGCN)} & \textcolor{black}{\textbf{0.3670 }} & \textcolor{black}{\textbf{0.7027 }} & \textcolor{black}{\textbf{0.6378 }} & \textcolor{black}{\textbf{0.4297 }} & \textcolor{black}{\underline{0.5903} } & \textcolor{black}{\textbf{0.5933 }} & \textcolor{black}{\textbf{0.5300 }} & \textcolor{black}{\underline{0.3743} } & \textcolor{black}{\underline{0.5300} } \\
    \bottomrule
    \end{tabular}%
  \label{tab:compare}%
  \vspace{-0.3cm}
\end{table*}%

\vspace{0.1cm}
\subsubsection{Implementation Details}
We use Adam \cite{kingma2014adam} optimizer to minimize the overall loss $\mathcal{L}$ of MDGCN and MSE loss $\mathcal{L}_{M}$ of DDIGCN. The learning rates used to optimize MDGCN and DDIGCN are 0.01 and 0.001, respectively. The training epochs for MDGCN and DDIGCN are set to 1000 and 400. The hidden representation size is fixed to 64. For MDGCN, LeakyReLU \cite{glorot2011deep} activation is used after the fully connected layers. The size of the graph convolution layer we set for MDGCN is 2. We set the hyperparameters $\beta_t=1/(t+2)$ and $\delta=1$. For DDIGCN, we set the layer sizes as 3 for the graph convolution. Batch normalization \cite{ioffe2015batch} and ReLU \cite{glorot2011deep} activation are applied after each layer.

\subsection{Experimental Results of Medication Suggestion}\label{exp_result}
% For a suggestion with $k$ drugs, we evaluate the effectiveness of these drugs when $k$ is taken from 1 to 6. 
The experimental results for medication suggestions are shown in Table~\ref{tab:compare}.
% listed in Table \ref{tab:topkrec}. 
We observe that DSSDDI achieves the best results for almost all $k$ values in $1,\ldots, 6$.
% , especially when $k\geq 5$, Recall@$k$ reaches more than 80\% and NDCG@$k$ reaches around 70\% on the test set. 
When compared with the graph learning-based methods, we find that traditional methods perform much worse. This is because traditional methods cannot capture latent patient features, making them difficult to provide effective suggestions based only on the patients' numerical features.
% when patients have similar numerical features, it is difficult for traditional methods to distinguish the differences between patients.
\eat{
In addition, we observe that DSSDDI works better than the graph learning-based methods.
%As we illustrate in Section \ref{Personalized} that this is due to the graph convolution process smooths the patient representations.
% we conclude that whether or not updating the patient representations through the graph convolution layer significantly affects the effectiveness of medication suggestions. 
This is due to the graph convolution process smooths the patient representations.
Since DSSDDI employs the patient representations before the graph convolution layer to predict the patient's medication use, which ensures that the patient representations during model training will not be effected by the similar drug representations. However, in LightGCN and GCMC, patient representations are updated through the graph convolution, resulting in most patients have similar representations. 
Nevertheless, LightGCN performs better than GCMC, which is because LightGCN eliminates feature transformation and activation function, and adds patient representations from previous layers after each graph convolution layer, resulting in less feature space changes affected by the graph convolution layers. 
}

\begin{figure*}[h]
% \vspace{-0.2cm}
\centering
\subfigure[Cosine similarity between patient representations.]{
\label{patientRepr}
\begin{minipage}{0.45\linewidth}
% \centering
\includegraphics[width=0.95\columnwidth]{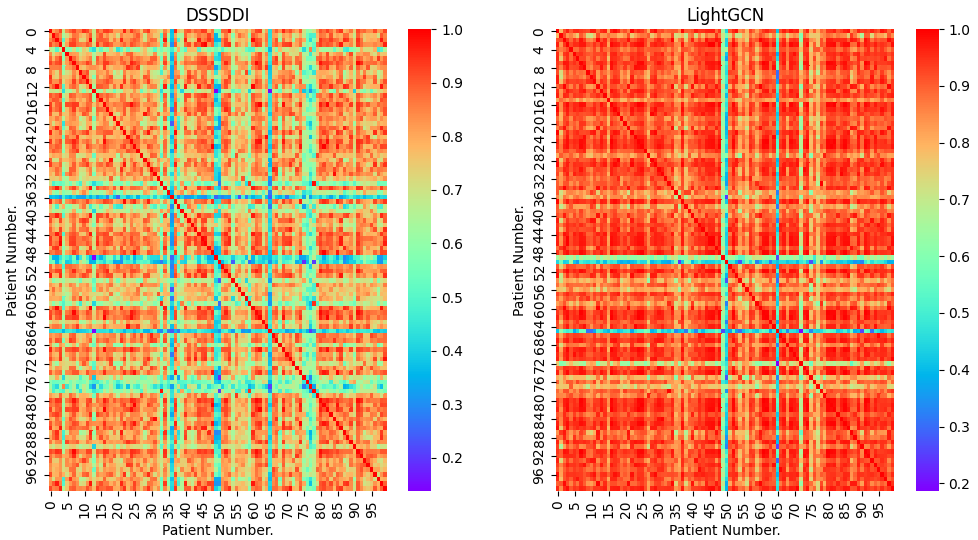}
\end{minipage}
}
\subfigure[Cosine similarity between drug representations.]{
\label{drugRepr}
\begin{minipage}{0.45\linewidth}
% \centering
\includegraphics[width=0.95\columnwidth]{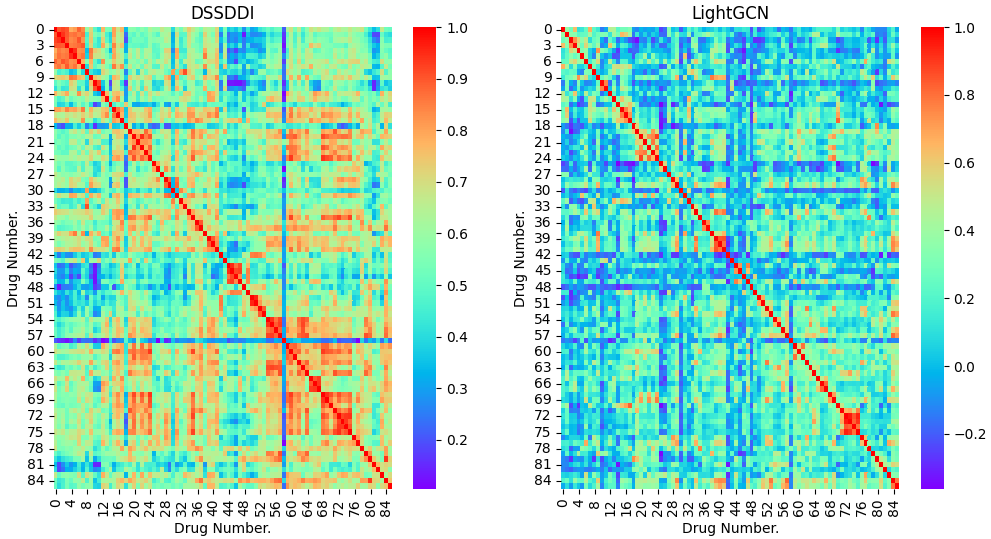}
\end{minipage}
}
\vspace{-0.3cm}
\caption{Comparison of cosine similarity between patient (drug) representations obtained from DSSDDI and LightGCN. Closer to red means that the representations are more similar and closer to blue means that they are less similar. (a) We sample 100 patients in the test set of patients. (b) 86 drugs are included in the comparison of their similarity to each other.}
\vspace{-0.4cm}
\label{heatmap}
\end{figure*}

In addition, we observe that DSSDDI performs better than the graph learning-based methods. The reason is that the graph convolution process smooths the patient representations. Specifically, DSSDDI employs the patient representations before the graph convolution layer to predict the patient's medication use, which ensures that the patient representations during model training will not be affected by similar drug representations. However, in graph learning-based methods, patient representations are updated through graph convolution, resulting in similar representations of most patients. To validate this speculation, we sample 100 patients in the test set and calculate the cosine similarity between patient representations by DSSDDI and LightGCN and plot the heat map in Fig. \ref{patientRepr}. As we can see, the similarity between the patient representations obtained by LightGCN is extremely high. Such similar representations make it difficult for LightGCN to identify feature differences between patients. In contrast, the patient representations obtained by DSSDDI are more distinguishable, which indicates that MDGCN avoids the over-smoothing of patient representations. 
Due to the high similarity between patient representations, some drugs with high similarity to patient representations will be suggested frequently, making it difficult for LightGCN to personalize the suggestions.

In addition, we calculate the cosine similarity between drug representations by DSSDDI and LightGCN and plot the heat map in Fig. \ref{drugRepr}. The drug representations obtained by DSSDDI are more reasonable because many drugs are related as they treat the same type of disease such as cardiovascular disease, arthritis and diabetes as shown in Fig \ref{drug_dis_Prec}. In contrast, all drug representations learned by LightGCN have low similarity.

\btrevise{When comparing DSSDDI with different backbones, DSSDDI(SGCN) performs best, showing that SGCN can learn better drug embeddings from synergistic and antagonistic interactions. DSSDDI(GIN) performs slightly worse than DSSDDI(SGCN). The attention-based models DSSDDI(SiGAT) and DSSDDI(SNEA) are less effective.}

\vspace{-0.2cm}
\begin{table*}[htbp]
\vspace{-0.2cm}
  \centering
  \caption{Ablation studies with different drug embeddings on Chronic data set (the best results are in bold). Here, we use SGCN, the best performing backbone model in Table~\ref{tab:compare}, as the backbone in DDIGCN.}
\vspace{-0.2cm}
   \begin{tabular}{llllllllll}
    \toprule
    \textcolor{black}{Method} & \textcolor{black}{Precision@6} & \textcolor{black}{Recall@6} & \textcolor{black}{NDCG@6} & \textcolor{black}{Precision@5} & \textcolor{black}{Recall@5} & \textcolor{black}{NDCG@5} & \textcolor{black}{Precision@4} & \textcolor{black}{Recall@4} & \textcolor{black}{NDCG@4} \\
    \midrule
    \textcolor{black}{w/o DDI} & \textcolor{black}{0.2185 } & \textcolor{black}{0.7974 } & \textcolor{black}{0.6427 } & \textcolor{black}{0.2490 } & \textcolor{black}{0.7694 } & \textcolor{black}{0.6301 } & \textcolor{black}{0.2891 } & \textcolor{black}{0.7256 } & \textcolor{black}{0.6089 } \\
    \textcolor{black}{One-hot} & \textcolor{black}{0.2095 } & \textcolor{black}{0.7952 } & \textcolor{black}{0.6063 } & \textcolor{black}{0.2365 } & \textcolor{black}{0.7537 } & \textcolor{black}{0.5891 } & \textcolor{black}{0.2638 } & \textcolor{black}{0.6830 } & \textcolor{black}{0.5574 } \\
    \textcolor{black}{KG} & \textcolor{black}{0.2135 } & \textcolor{black}{0.8170 } & \textcolor{black}{0.6489 } & \textcolor{black}{0.2411 } & \textcolor{black}{0.7761 } & \textcolor{black}{0.6319 } & \textcolor{black}{0.2758 } & \textcolor{black}{0.7187 } & \textcolor{black}{0.6067 } \\
    % \midrule
    \textcolor{black}{DDIGCN} & \textcolor{black}{\textbf{0.2348 }} & \textcolor{black}{\textbf{0.8521 }} & \textcolor{black}{\textbf{0.6850 }} & \textcolor{black}{\textbf{0.2670 }} & \textcolor{black}{\textbf{0.8153 }} & \textcolor{black}{\textbf{0.6717 }} & \textcolor{black}{\textbf{0.3077 }} & \textcolor{black}{\textbf{0.7746 }} & \textcolor{black}{\textbf{0.6680 }} \\
    \midrule
    \midrule
    \textcolor{black}{Method} & \textcolor{black}{Precision@3} & \textcolor{black}{Recall@3} & \textcolor{black}{NDCG@3} & \textcolor{black}{Precision@2} & \textcolor{black}{Recall@2} & \textcolor{black}{NDCG@2} & \textcolor{black}{Precision@1} & \textcolor{black}{Recall@1} & \textcolor{black}{NDCG@1} \\
    \midrule
    \textcolor{black}{w/o DDI} & \textcolor{black}{0.3413} & \textcolor{black}{0.6499} & \textcolor{black}{0.5788} & \textcolor{black}{0.4032} & \textcolor{black}{0.5277} & \textcolor{black}{0.5292} & \textcolor{black}{0.4796} & \textcolor{black}{0.3204} & \textcolor{black}{0.4796} \\
    \textcolor{black}{One-hot} & \textcolor{black}{0.2984 } & \textcolor{black}{0.5904 } & \textcolor{black}{0.5150 } & \textcolor{black}{0.3462 } & \textcolor{black}{0.4715 } & \textcolor{black}{0.4635 } & \textcolor{black}{0.4147 } & \textcolor{black}{0.2855 } & \textcolor{black}{0.4147 } \\
    \textcolor{black}{KG} & \textcolor{black}{0.3297 } & \textcolor{black}{0.6609 } & \textcolor{black}{0.5810 } & \textcolor{black}{0.3918 } & \textcolor{black}{0.5425 } & \textcolor{black}{0.5304 } & \textcolor{black}{0.4591 } & \textcolor{black}{0.3338 } & \textcolor{black}{0.4591 } \\
    % \midrule
    \textcolor{black}{DDIGCN} & \textcolor{black}{\textbf{0.3670 }} & \textcolor{black}{\textbf{0.7027 }} & \textcolor{black}{\textbf{0.6378 }} & \textcolor{black}{\textbf{0.4297 }} & \textcolor{black}{\textbf{0.5903 }} & \textcolor{black}{\textbf{0.5933 }} & \textcolor{black}{\textbf{0.5300 }} & \textcolor{black}{\textbf{0.3743 }} & \textcolor{black}{\textbf{0.5300 }} \\
    \bottomrule
    \end{tabular}%
  \label{tab:abstudy}%
  \vspace{-0.2cm}
\end{table*}%

\vspace{-0.5cm}
\btrevise{
\subsection{Superiority of DDIGCN}\label{SecCase}
We conduct an ablation study to demonstrate the superiority of DDIGCN. We create the following variants by replacing the drug embedding learned by DDIGCN with three alternatives:
\begin{itemize}
\item \textbf{Without DDI}: Without adding DDI relation embeddings to the final drug embeddings $\textbf{h}'_v$.
\item \textbf{One-hot}: One-hot embeddings.
\item \textbf{KG}: Pre-trained embeddings obtained from DRKG~\cite{drkg2020}.
\end{itemize}
}
\btrevise{
The results are reported in Table~\ref{tab:abstudy}. DDIGCN performs the best, which indicates that DDIGCN learns drug embeddings from the DDI relationship that is more useful for medication suggestions. DDIGCN outperforms the system without DDI module, due to its consideration of drug-drug interactions, thus avoiding numerous inappropriate suggestions.  As DRKG contains many different types of entities such as genes and proteins, KG per-trained embeddings may contain much too complex information, such as the relationship between drug and genes which affect the drug suggestion performance.}

\begin{table}[t]
\vspace{-0.2cm}
\setlength\tabcolsep{5pt}
  \centering
  \caption{Suggestion satisfaction comparison between the proposed method and baseline methods on $k$ medication suggestions.}
  \vspace{-0.2cm}
    \begin{tabular}{lllllll}
    \toprule
    k     & 2     & 3     & 4     & 5     & 6  \\
    \midrule
    \midrule
    UserSim & 0.4987  & 0.2506  & 0.0743  & 0.0470  & 0.0220  \\
    ECC   & 0.5000  & 0.2500  & 0.0952  & 0.0455  & 0.0339  \\
    SVM   & 0.5044  & 0.2695  & 0.1050  & 0.0469  & 0.0278  \\
    \midrule
    GCMC  & 0.4979  & 0.2533  & 0.1443  & 0.0491  & 0.0634  \\
    LightGCN & 0.5046  & 0.2544  & 0.1631  & 0.0882  & 0.0575  \\
    \textcolor{black}{SafeDrug} & \textcolor{black}{0.4412 } & \textcolor{black}{0.1812 } & \textcolor{black}{0.0741 } & \textcolor{black}{0.0477 } & \textcolor{black}{0.0329 } \\
    \textcolor{black}{Bipar-GCN} & \textcolor{black}{0.5139 } & \textcolor{black}{0.2788 } & \textcolor{black}{0.1735 } & \textcolor{black}{0.1183 } & \textcolor{black}{0.0866 } \\
    \textcolor{black}{CauseRec} & \textcolor{black}{0.4957 } & \textcolor{black}{0.2462 } & \textcolor{black}{0.0996 } & \textcolor{black}{0.0482 } & \textcolor{black}{0.0299 } \\
    \midrule
    \textcolor{black}{DSSDDI(SiGAT)} & \textcolor{black}{\textbf{0.5683 }} & \textcolor{black}{0.3237 } & \textcolor{black}{0.2043 } & \textcolor{black}{0.1400 } & \textcolor{black}{0.1042 } \\
    \textcolor{black}{DSSDDI(SNEA)} & \textcolor{black}{0.5522 } & \textcolor{black}{0.2916 } & \textcolor{black}{0.1811 } & \textcolor{black}{0.1253 } & \textcolor{black}{0.0899 } \\
    DSSDDI(GIN) & 0.5392  & 0.2767  & 0.1743  & 0.1227  & 0.0997  \\
    \textcolor{black}{DSSDDI(SGCN)} & \textcolor{black}{0.5427 } & \textcolor{black}{\textbf{0.3267 }} & \textcolor{black}{\textbf{0.2158 }} & \textcolor{black}{\textbf{0.1478 }} & \textcolor{black}{\textbf{0.1083 }} \\
    \bottomrule
    \end{tabular}%
  \label{tab:ss}%
  \vspace{-0.4cm}
\end{table}%

\subsection{Explainability of DSSDDI}
Table \ref{tab:ss} compares the proposed method with baseline methods in terms of \textit{Suggestion Satisfaction (SS)} with medication suggestions.
% \rr{It seem that we need some numbers here. e.g., Compared with the bestbaselines, DSSDDI imporve the SS@k by XXX\%.}
DSSDDI has a significant improvement compared to other methods. In particular, compared with the best-performing baseline, DSSDDI improves $SS@6$ by \btrevise{25\%}, $SS@5$ by \btrevise{24\% and $SS@4$ by 24\%}. Because \textit{SS} takes into account not only the synergy between the $k$ suggested drugs but also the antagonism between the suggested and non-suggested drugs, we can conclude that DSSDDI is able to suggest drugs with more synergistic effects, while being able to avoid drugs with antagonistic effects from being included. By extracting these relevant drug-drug interactions as subgraphs in the DDI graph, it provides an effective explanation for the drugs suggested by the Medical Decision module.

\begin{figure*}[t]
% \vspace{-0.2cm}
\centering
\subfigure[DSSDDI]{
\label{fig.our}
\centering
\includegraphics[width=0.36\columnwidth]{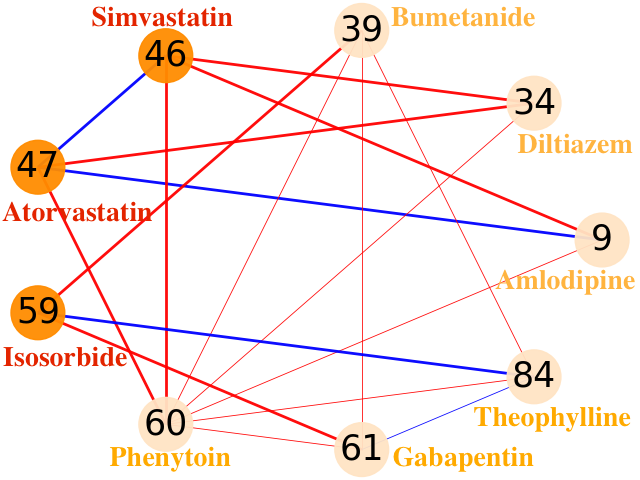}
}
\subfigure[LightGCN]{
\label{fig.lightgcn}
\centering
\includegraphics[width=0.36\columnwidth]{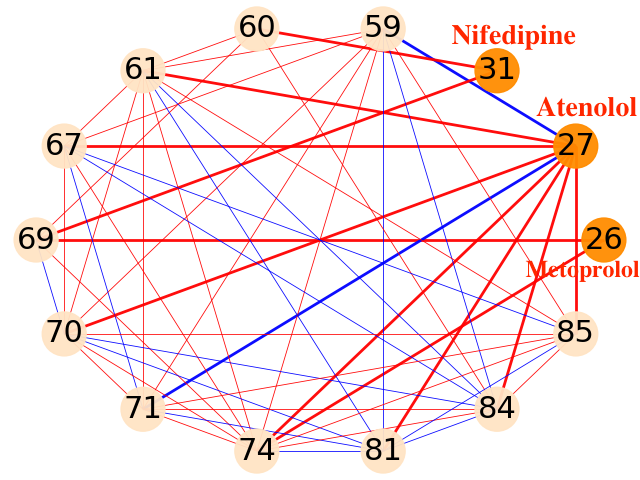}
}
\subfigure[GCMC]{
\label{fig.GCMC}
\centering
\includegraphics[width=0.36\columnwidth]{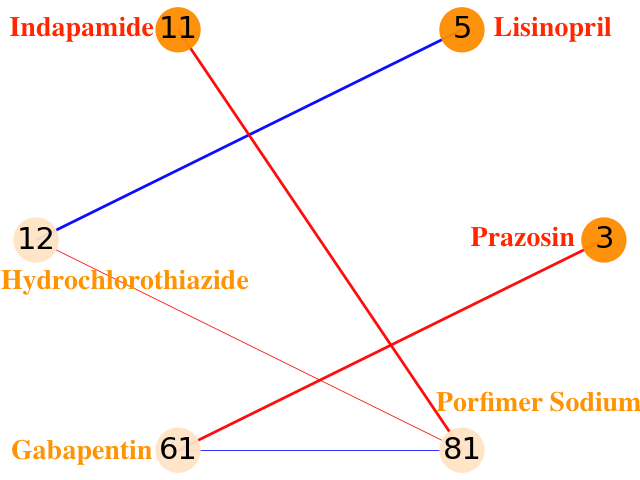}
}
\subfigure[SVM]{
\label{fig.SVM}
\centering
\includegraphics[width=0.36\columnwidth]{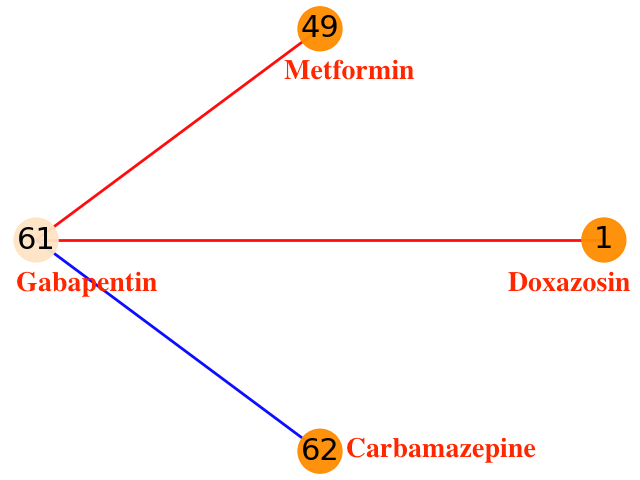}
}
\subfigure[ECC]{
\label{fig.ecc}
\centering
\includegraphics[width=0.36\columnwidth]{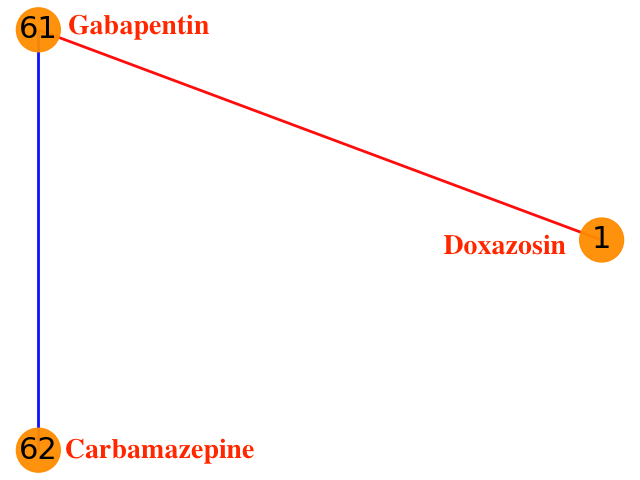}
}
\vspace{-0.25cm}
\caption{A case of medication suggestion for a patient with cardiovascular disease. We use dark orange nodes to denote suggested drugs and light orange nodes to denote non-suggested nodes in the subgraph output by the MS module, red lines to indicate the antagonism between drugs, and blue lines to indicate the synergy between drugs. We make the edges between the non-suggested drugs transparent to highlight the interactions associated with the suggested drugs.}
\label{fig.case}
\end{figure*}

We then show a case of medication suggestion for a patient with cardiovascular disease in Fig. \ref{fig.case} to highlight the more reliable medication suggestions of DSSDDI compared to baseline methods from an explainable perspective. 
All these subgraphs were obtained according to the subgraph querying algorithm proposed in the MS module.
As shown in Fig. \ref{fig.our}, DSSDDI suggests Simvastatin (Drug ID (DID) 46),  Atorvastatin (DID 47) and Isosorbide (DID 59). There is a synergistic effect between Simvastatin (DID 46) and Atorvastatin (DID 47) as indicated by the blue line between them. Using these two drugs is indeed beneficial in lowering lipids and improving cardiovascular disease. The suggestion by DSSDDI can also avoid antagonistic effects. For example, it does not suggest Gabapentin (DID 61) because there is an antagonistic effect between Gabapentin (DID 61) and Isosorbide (DID 59) as indicated by the red line between them. This shows that DSSDDI considers synergistic effects between the suggested drugs and avoids antagonism benefiting from the combination.

In contrast, the three drugs suggested by LightGCN, GCMC and SVM, respectively (in Figs. \ref{fig.lightgcn}-\ref{fig.SVM}) do not have any interactions. The drugs suggested by ECC even have antagonistic interactions, i.e., Gabapentin (DID 61), a drug for epilepsy, is antagonistic to Doxazosin (DID 1) as shown in Fig. \ref{fig.ecc}.

%LightGCN, GCMC, and SVM suggest drugs without interactions, and ECC even suggests drugs that include antagonistic interactions.
%In contrast, the drugs suggested by DSSDDI not only avoid antagonism but also consider synergistic effects between drugs benefiting from the combination with DDI. We analyze these subgraphs in terms of both synergistic and antagonistic effects as follows.
%\paragraph{Promoting synergistic effects} As shown in \ref{fig.our}, the DSSDDI suggests Simvastatin (Drug ID (DID) 46) and Atorvastatin (DID 47) together, which can indeed be beneficial in lowering lipids and improving cardiovascular disease, whereas other approaches ignore the synergistic effect when suggesting drugs. Such promotions provide doctors with more suitable alternative.
%\paragraph{Avoiding antagonistic effects} The ECC suggests Gabapentin (DID 61), a drug for epilepsy, which is apparently antagonistic to Doxazosin (DID 1) as shown in \ref{fig.ecc}. DSSDDI, however, does not suggest Gabapentin (DID 61) because it avoids the antagonistic effect between Isosorbide (DID 59) and Gabapentin (DID 61).

From this experiment, we can see that DSSDDI outputs such explainable DDI subgraphs through the Medical Support module, which can provide more convincing support for DSSDDI's medical suggestions and make doctors' decisions more efficient.
%这些子图是根据各方法建议的药物输入ED模块中提出的subgraphquerying算法得到的。
%从图8中可以看到，多数方法如LightGCN,GCMC,SVM建议的药物之间都不存在相互作用，甚至ECC所建议的药物之间还存在拮抗作用。
%从协同作用的角度看，DSSDDI所建议的药物，促进了药物之间可能存在的协同作用，如图中所示46 Simvastatin和47 Atorvastatin一起服用能有利于降低血脂，改善心血管疾病 [?]。
%从拮抗作用的角度看，ECC建议的61 Gabapentin是一种治疗癫痫的药物 显然与DSSDDI建议的59 Isosorbide和ECC自己建议的1Doxazosin之间有拮抗作用，DSSDDI通过59和61的拮抗作用避免了此类药物建议，而ECC因为缺乏DDI的知识产生了含有拮抗作用的建议。
%DSSDDI通过medical support component输出此类可解释性子图，能够为DSSDDI的medical建议提供更多的支撑，让医生的决策更高效。

\begin{figure*}[t]
  \centering
  \includegraphics[width=\linewidth]{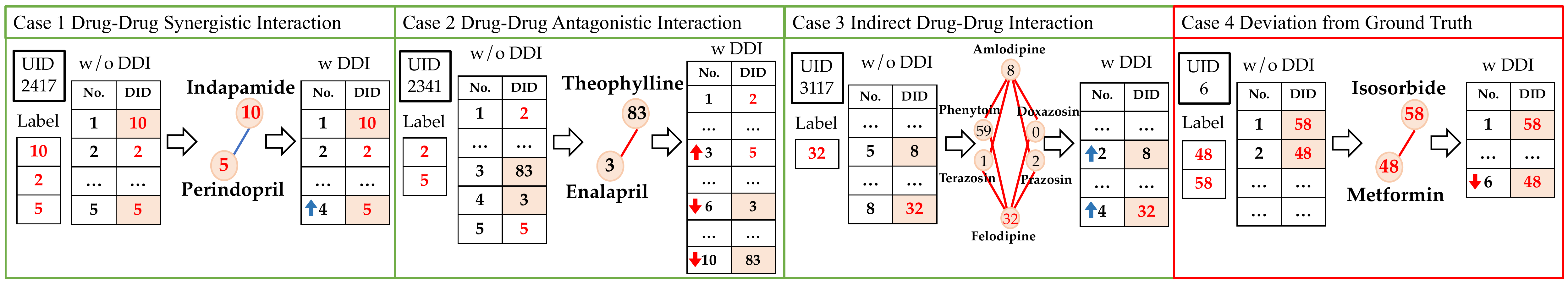}
  \vspace{-0.8cm}
  \caption{Four case studies that demonstrate the superiority of DDI. The red numbers in the Label column indicate the drugs the patient took, the blue (red) line indicates the drug-drug synergistic (antagonistic) effect, the blue (red) arrow indicates an upward or downward movement brought by the synergistic (antagonistic) effect.}
  \vspace{-0.2cm}
  \label{cases}
\end{figure*}

\begin{table*}[htbp]
\vspace{-0.1cm}
  \centering
    \caption{Medication suggestion performance comparison between the proposed method and baseline methods on MIMIC-III data set \newline (the best results are in bold).}
    \vspace{-0.2cm}
    \begin{tabular}{llllllllll}
    \toprule
    \textcolor{black}{Method} & \textcolor{black}{Precision@8} & \textcolor{black}{Recall@8} & \textcolor{black}{NDCG@8} & \textcolor{black}{Precision@6} & \textcolor{black}{Recall@6} & \textcolor{black}{NDCG@6} & \textcolor{black}{Precision@4} & \textcolor{black}{Recall@4} & \textcolor{black}{NDCG@4} \\
    \midrule
    \textcolor{black}{UserSim} & \textcolor{black}{0.5396 } & \textcolor{black}{0.2349 } & \textcolor{black}{0.6203 } & \textcolor{black}{0.5699 } & \textcolor{black}{0.1869 } & \textcolor{black}{0.6534 } & \textcolor{black}{0.7006 } & \textcolor{black}{0.1551 } & \textcolor{black}{0.7557 } \\
    \textcolor{black}{ECC} & \textcolor{black}{0.6957 } & \textcolor{black}{0.3014 } & \textcolor{black}{0.7360 } & \textcolor{black}{0.7786 } & \textcolor{black}{0.2562 } & \textcolor{black}{0.7904 } & \textcolor{black}{0.8116 } & \textcolor{black}{0.1795 } & \textcolor{black}{0.8111 } \\
    \textcolor{black}{SVM} & \textcolor{black}{0.7645 } & \textcolor{black}{0.3343 } & \textcolor{black}{0.7829 } & \textcolor{black}{0.8234 } & \textcolor{black}{0.2703 } & \textcolor{black}{0.8182 } & \textcolor{black}{0.8313 } & \textcolor{black}{0.1833 } & \textcolor{black}{0.8210 } \\
    \midrule
    \textcolor{black}{GCMC} & \textcolor{black}{0.7924 } & \textcolor{black}{0.3283 } & \textcolor{black}{0.8156 } & \textcolor{black}{0.8186 } & \textcolor{black}{0.2656 } & \textcolor{black}{0.8208 } & \textcolor{black}{0.8360 } & \textcolor{black}{0.1764 } & \textcolor{black}{0.8392 } \\
    \textcolor{black}{LightGCN} & \textcolor{black}{0.8099 } & \textcolor{black}{0.3548 } & \textcolor{black}{0.8252 } & \textcolor{black}{0.8310 } & \textcolor{black}{0.2738 } & \textcolor{black}{0.8378 } & \textcolor{black}{0.8449 } & \textcolor{black}{0.1872 } & \textcolor{black}{0.8495 } \\
    \textcolor{black}{SafeDrug} & \textcolor{black}{0.8038 } & \textcolor{black}{0.3549 } & \textcolor{black}{0.8215 } & \textcolor{black}{0.8172 } & \textcolor{black}{0.2701 } & \textcolor{black}{0.8308 } & \textcolor{black}{0.8434 } & \textcolor{black}{0.1869 } & \textcolor{black}{0.8494 } \\
    \textcolor{black}{Bipar-GCN} & \textcolor{black}{0.7939 } & \textcolor{black}{0.3510 } & \textcolor{black}{0.8135 } & \textcolor{black}{0.8172 } & \textcolor{black}{0.2718 } & \textcolor{black}{0.8281 } & \textcolor{black}{0.8327 } & \textcolor{black}{0.1856 } & \textcolor{black}{0.8390 } \\
    \textcolor{black}{CauseRec} & \textcolor{black}{0.1218 } & \textcolor{black}{0.1428 } & \textcolor{black}{0.1346 } & \textcolor{black}{0.1196 } & \textcolor{black}{0.1130 } & \textcolor{black}{0.1238 } & \textcolor{black}{0.1157 } & \textcolor{black}{0.0818 } & \textcolor{black}{0.1162 } \\
    \midrule
    \textcolor{black}{DSSDDI(GIN)} & \textcolor{black}{\textbf{0.8134 }} & \textcolor{black}{\textbf{0.3611 }} & \textcolor{black}{\textbf{0.8266 }} & \textcolor{black}{\textbf{0.8352 }} & \textcolor{black}{\textbf{0.2808 }} & \textcolor{black}{\textbf{0.8408 }} & \textcolor{black}{\textbf{0.8530 }} & \textcolor{black}{\textbf{0.1932 }} & \textcolor{black}{\textbf{0.8553 }} \\
    \bottomrule
    \end{tabular}%
  \label{tab:mimic}%
  \vspace{-0.3cm}
\end{table*}%

\vspace{-0.6cm}
\btrevise{\subsection{Validation on Public Diagnostic Data}
In this experiment, we evaluate medication suggestion performance using a public, real diagnostic data set, MIMIC-III~\cite{johnson2016mimic}. It is a database comprising de-identified health-related data associated with patients who stayed in critical care units of the Beth Israel Deaconess Medical Center between 2001 and 2012. 
% Based on the processed data\footnote{https://github.com/ycq091044/SafeDrug}, we obtain 
This data set includes 6350 patients, each with at least two visits. Each visit includes three types of information: medicine, diagnosis and procedure. To fit our problem setting, we use the medicine information of the last visit as the medication suggestion label to be predicted, and the diagnosis and procedure information of previous visits as the patient features. The medication suggestion performance of the proposed method and the baselines is shown in Table~\ref{tab:mimic}.  Since only antagonistic interactions between anonymous drugs are included in the downloaded data, we cannot use signed graph-based models as the backbone, so only GIN-based results are reported in Table~\ref{tab:mimic}. DSSDDI outperforms all baselines, demonstrating the significant effects that the DDI brings to medical suggestions.
}

\vspace{0.1cm}
\section{System Application}
%As shown in Fig. 1, we deploy the proposed decision support system to help doctors provide medication suggestions to patients with chronic diseases. The background and data collection process are introduced in Section \ref{datacol}. 
In this section, we present four cases to further illustrate the superiority of DDI in Fig. \ref{cases}. 
% \subsection{Deployment and Application}
%we first show some cases that demonstrate the direct enhancement by the DDI module, and then analyze a case in which the DDI module indirectly affects medical decision. Finally, we present a case where the patient is taking drugs that contain antagonistic effects. In such cases DSSDDI has difficulty in achieving a better suggestion.

\noindent\textbf{Case 1}. The patient 2417 took medications DID 10, 2, 5 (in the ``Label'' column).  LightGCN without using DDI ranks DID 10, 2 and 5 at the first, second and fifth positions, respectively. As there is a synergistic interaction between Indapamide (DID 10) and Perindopril (DID 5), our model can rank Perindopril at the fourth position (instead of the fifth) due to this synergy. Thus our model provides a more accurate suggestion by using DDI information.

\noindent\textbf{Case 2}. The patient 2341 took medications DID 2 and 5. LightGCN without using DDI ranks DID 2 and 5 at the first and fifth positions, respectively. It also suggests Theophylline (DID 83) and Enalapril (DID 3) at the third and fourth position.  However, there is an antagonistic interaction between them, as indicated by the red line.  In contrast, our model ranks DID 2 and 5 at the first and third position, and ranks DID 3 and 83 at a lower position.  This suggestion is more accurate than that by w/o DDI. 

\noindent\textbf{Case 3}. The patient 3117 took medication DID 32. LightGCN w/o DDI ranks DID 32 at the eighth position. We find Amlodipine (DID 8) and Felodipine (DID 32) are antagonistic to four common drugs, including Phenytoin, Doxazosin, Terazosin, and Prazosin.  Therefore, these two are considered as similar drugs by DSSDDI. Indeed both of them are for treating hypertension.  Benefiting from the message passing of DDIGCN, these two drugs can have similar representations due to interactions with many common drugs, although they do not have direct drug-drug interactions. As a result, our model ranks these two drugs at the second and fourth positions, respectively. 

\noindent\textbf{Case 4}. In practice, we observe that some patients are taking some drugs with antagonistic effects, probably because they were facing severe medical conditions, thus had to neglect the drug-drug interactions. For instance, the patient 6 took Isosorbide (DID 58) and Metformin (DID 48) which may cause adverse drug reactions such as cholecystitis and dizziness, so DSSDDI downgrades Metformin to the sixth position.  Although this suggestion is not consistent with the ground truth, it appears more reasonable from the perspective of DDI due to the antagonistic effect between the two drugs.

\section{Related work}\label{review}
This work is related to decision support system, graph learning-based recommendation systems, causal models with graph learning and Drug-Drug Interaction models.

\subsection{Decision Support System}
\vspace{-0.1cm}
Decision Support System is a computer software that provides a reference for medical practitioners in their clinic decision-making \cite{osheroff2007roadmap}. Recent decision support systems can be divided into knowledge-based and data-driven methods \cite{awaysheh2019review}. 

Knowledge-based methods are mainly developed based on medical guidelines and medical knowledge, for example, rule-based methods \cite{segundo2017improvement}, semantic or associative networks \cite{seol2017causality} and Bayesian Networks \cite{elkin2018artificial}. These methods are generally limited in scale, due to the lack of evidence in some domains.

With the development of Machine Learning (ML), more and more decision support systems are shifting from knowledge-based methods to data-driven methods~\cite{awaysheh2019review}. ML models such as \btrevise{Ensemble Classifier Chain (ECC)~\cite{read2009classifier}, Support Vector Machines (SVM)~\cite{ma2016first} have been widely applied to develop decision support systems. However, they rely on the quality and quantity of data provided. When there is a bias in the data used to train the ML model, this bias is captured by the model and therefore biased predictions are made which can influence human decisions.} Therefore, the aim of this paper is to propose an explainable decision support system that can provide explanations for the suggested medication, which will be more beneficial for medical practitioners to make decisions.

\subsection{Graph Learning-based Recommendation Systems}
\vspace{-0.1cm}
Since the data in most recommendation systems is essentially a graph structure, more and more graph learning methods have been applied to learn the inter-object relations in recommendation systems \cite{li2020graph}. 
% Considering the graph where the objects ( e.g., users, items) are nodes and relations (e.g., social relations, interaction history) between them are edges, graph learning based recommendation systems predict the corresponding recommendation results by modeling the topological structure and content information on the graph. 
Random walk based recommendation system \cite{baluja2008video} has been widely adopted to capture complex, higher-order and indirect relations among a variety of nodes on the graph \cite{eksombatchai2018pixie}. 
% These methods use a random walker to walk on the graph and then rank the candidate nodes for recommendation based on the probability that the walker lands on those nodes after certain steps. 
% Since these random walk based recommendation systems need to generate ranking scores for all candidate nodes at each step, and thus are difficult to apply on large-scale graphs because of the low efficiency.
Graph representation learning-based recommendation system~\cite{wang2019unified} encodes each node into a latent representation and then analyzes the complex relations between them. Another class of recommendation systems are built on the knowledge graph (KG) to explore latent relations among users or items connected as a heterogeneous graph\cite{wang2018dkn}.
% Methods such as  matrix factorization \cite{wang2019unified}, skip-gram model \cite{mikolov2013distributed} and neural networks \cite{cen2019representation} are employed to obtain the latent representations of each node. 

Benefiting from the strength of Graph Neural Networks (GNN), a lot of GNN based recommendation systems \cite{li2021deconvolutional, he2020lightgcn} are developed. \btrevise{GCMC~\cite{berg2017graph} leverages matrix completion to obtain the latent node representations. LightGCN~\cite{he2020lightgcn} simplifies the graph convolution by eliminating the nonlinear activation function and feature transformation matrix. Bipar-GCN~\cite{DBLP:conf/icde/JinZ00W20} learns user representations and drug representations by training user-oriented and item-oriented neural networks respectively. However, these methods have the problem of over-smoothing of patient representations because many patients take similar medications. The proposed DSSDDI solves this problem by using the patient representations before the graph convolutional operation.} \btrevise{SafeDrug~\cite{DBLP:conf/ijcai/YangXMGS21} combines the drug molecular graph and DDI graph to predict safe medication combinations. However, this method is difficult for new patients because it relies on medication information from patient's past visits to generate patient features.} 

% Our decision support system differs from the KG recommendation system, it is a process of suggesting medications for patients, which are connected as a bipartite graph.

\subsection{Causal Models with Graph Learning}
\vspace{-0.1cm}
As causal models capture the causal relationship between outcomes and inputs, a series of studies have focused on using causal models to enhance the inference stage of ML models \cite{hu2021distilling,yang2021deconfounded,chang2013know}. 
Recently, several causal models with graph learning have gained increasing attention \cite{feng2021should,lin2021generative,bevilacqua2021size}. %For example, Feng \textit{et al.} \cite{feng2021should} compare the original prediction with the intervened prediction to assess the causal effect of the local structure
%on the node prediction. Bevilacqua \textit{et al.} solve out-of-distribution (OOD) graph classification by employing a causal model to learn approximately invariant representations that better extrapolate between training and test data. 
Based on causal models, counterfactual causal inference \cite{morgan2015counterfactuals} aims to find out the causal relationship between treatments and outcomes by exploring whether the outcomes would be different if the treatment is different. Zhao \textit{et al.} \cite{zhao2022learning} employ the counterfactual causal model to improve graph representation learning. \btrevise{CauseRec~\cite{DBLP:conf/sigir/ZhangYZC021} generates counterfactual patient behavior sequences to learn patient representations. However, it relies mainly on medication information from
patients’ past visits, making it difficult to cope with many new patients on their first visit.} In this paper, we explore the application of the counterfactual causal model to medical decision.

\subsection{Drug-Drug Interaction Models}
\vspace{-0.1cm}
Recently, a lot of methods have been proposed for learning low-dimensional drug embeddings which are then used for downstream tasks such as Drug-Drug Interaction (DDI) prediction or drug classification \cite{yu2022raneddi}. These approaches can be mainly classified into relational-based approaches and network structure-based approaches. 

Relational-based approaches use knowledge graph to capture multi-relational information from different edge relations \cite{lin2015learning}. For example, Zitnik \textit{et al.} \cite{zitnik2016collective} employ a multi-relational network to identify the relation between drugs by regrading each relation as a matrix. TransE \cite{bordes2013translating} and TransH \cite{wang2014knowledge} are also commonly used knowledge representation learning models being used to learn drug embeddings and the relations \cite{drkg2020}. 

Unlike the relational-based approaches, network structure-based approaches employ Graph Neural Networks to learn the drug embeddings through aggregating the information of their neighbors \cite{wu2020comprehensive}. For instance, Feng \textit{et al.} \cite{feng2020dpddi} apply GCN to capture the network structure information for drugs in the DDI network. Wang \textit{et al.} \cite{wang2020gognn} develop a dual-attention model to extract features from the molecular and the DDI graphs.

\section{Conclusion}\label{conclu}
In this paper, we design a decision support system called DSSDDI for assisting doctors in making clinical decisions for patients with chronic diseases. Based on the external knowledge of drug-drug interactions, DSSDDI is able to obtain not only reliable medication suggestions through causal relationships in the Medical Decision module, but also the explanations for the suggestions through the Medical Support module. DSSDDI achieves superior performance to the baseline approaches for medication suggestion. In addition, it also provides explanation to the suggestions. In future work, we plan to complement the decision support system with some medical image features and to enable the search for etiological and explainable factors on the images.

\section*{Acknowledgment}
This research is funded by National Key R\&D Program of China 2018YFC2000702 and partially supported by the Stanley Ho Big Data Decision Analytics Research Centre at The Chinese University of Hong Kong. Jia Li is supported by NSFC Grant No. 62206067 and Tencent AI Lab Rhino-Bird Focused Research Program RBFR2022008.

\newpage

\bibliographystyle{IEEEtran.bst}
\bibliography{IEEEexample.bib}

\begin{thebibliography}{10}
\providecommand{\url}[1]{#1}
\csname url@samestyle\endcsname
\providecommand{\newblock}{\relax}
\providecommand{\bibinfo}[2]{#2}
\providecommand{\BIBentrySTDinterwordspacing}{\spaceskip=0pt\relax}
\providecommand{\BIBentryALTinterwordstretchfactor}{4}
\providecommand{\BIBentryALTinterwordspacing}{\spaceskip=\fontdimen2\font plus
\BIBentryALTinterwordstretchfactor\fontdimen3\font minus
  \fontdimen4\font\relax}
\providecommand{\BIBforeignlanguage}[2]{{%
\expandafter\ifx\csname l@#1\endcsname\relax
\typeout{** WARNING: IEEEtran.bst: No hyphenation pattern has been}%
\typeout{** loaded for the language `#1'. Using the pattern for}%
\typeout{** the default language instead.}%
\else
\language=\csname l@#1\endcsname
\fi
#2}}
\providecommand{\BIBdecl}{\relax}
\BIBdecl

\bibitem{wimmer2017clinical}
B.~C. Wimmer, A.~J. Cross, N.~Jokanovic, M.~D. Wiese, J.~George, K.~Johnell,
  B.~Diug, and J.~S. Bell, ``Clinical outcomes associated with medication
  regimen complexity in older people: a systematic review,'' \emph{JAGS},
  vol.~65, no.~4, pp. 747--753, 2017.

\bibitem{dumbreck2015drug}
S.~Dumbreck, A.~Flynn, M.~Nairn, M.~Wilson, S.~Treweek, S.~W. Mercer,
  P.~Alderson, A.~Thompson, K.~Payne, and B.~Guthrie, ``Drug-disease and
  drug-drug interactions: systematic examination of recommendations in 12 uk
  national clinical guidelines,'' \emph{BMJ}, vol. 350, 2015.

\bibitem{awaysheh2019review}
A.~Awaysheh, J.~Wilcke, F.~Elvinger, L.~Rees, W.~Fan, and K.~L. Zimmerman,
  ``Review of medical decision support and machine-learning methods,''
  \emph{Veterinary pathology}, vol.~56, no.~4, pp. 512--525, 2019.

\bibitem{DBLP:conf/icde/JinZ00W20}
Y.~Jin, W.~Zhang, X.~He, X.~Wang, and X.~Wang, ``Syndrome-aware herb
  recommendation with multi-graph convolution network,'' in \emph{{ICDE}
  2020}.\hskip 1em plus 0.5em minus 0.4em\relax {IEEE}, pp. 145--156.

\bibitem{DBLP:conf/sigir/ZhangYZC021}
S.~Zhang, D.~Yao, Z.~Zhao, T.~Chua, and F.~Wu, ``Causerec: Counterfactual user
  sequence synthesis for sequential recommendation,'' in \emph{{SIGIR}
  '21}.\hskip 1em plus 0.5em minus 0.4em\relax {ACM}, 2021, pp. 367--377.

\bibitem{DBLP:conf/ijcai/YangXMGS21}
C.~Yang, C.~Xiao, F.~Ma, L.~Glass, and J.~Sun, ``Safedrug: Dual molecular graph
  encoders for recommending effective and safe drug combinations,'' in
  \emph{{IJCAI} 2021}, pp. 3735--3741.

\bibitem{hu2021distilling}
X.~Hu, K.~Tang, C.~Miao, X.-S. Hua, and H.~Zhang, ``Distilling causal effect of
  data in class-incremental learning,'' in \emph{CVPR}, 2021, pp. 3957--3966.

\bibitem{yang2021deconfounded}
X.~Yang, F.~Feng, W.~Ji, M.~Wang, and T.-S. Chua, ``Deconfounded video moment
  retrieval with causal intervention,'' in \emph{SIGIR}, 2021, pp. 1--10.

\bibitem{johnson2016mimic}
A.~E. Johnson, T.~J. Pollard, L.~Shen, L.-w.~H. Lehman, M.~Feng, M.~Ghassemi,
  B.~Moody, P.~Szolovits, L.~Anthony~Celi, and R.~G. Mark, ``Mimic-iii, a
  freely accessible critical care database,'' \emph{Scientific data}, vol.~3,
  no.~1, pp. 1--9, 2016.

\bibitem{drkg2020}
V.~N. Ioannidis, X.~Song, S.~Manchanda, M.~Li, X.~Pan, D.~Zheng, X.~Ning,
  X.~Zeng, and G.~Karypis, ``Drkg - drug repurposing knowledge graph for
  covid-19,'' 2020.

\bibitem{bordes2013translating}
A.~Bordes, N.~Usunier, A.~Garcia-Duran, J.~Weston, and O.~Yakhnenko,
  ``Translating embeddings for modeling multi-relational data,''
  \emph{NeurIPS}, vol.~26, 2013.

\bibitem{liu2020drugcombdb}
H.~Liu, W.~Zhang, B.~Zou, J.~Wang, Y.~Deng, and L.~Deng, ``Drugcombdb: a
  comprehensive database of drug combinations toward the discovery of
  combinatorial therapy,'' \emph{Nucleic acids research}, vol.~48, no.~D1, pp.
  D871--D881, 2020.

\bibitem{xu2018powerful}
K.~Xu, W.~Hu, J.~Leskovec, and S.~Jegelka, ``How powerful are graph neural
  networks?'' in \emph{ICLR}, 2018.

\bibitem{DBLP:conf/icdm/Derr0T18}
T.~Derr, Y.~Ma, and J.~Tang, ``Signed graph convolutional networks,'' in
  \emph{ICDM 2018}.\hskip 1em plus 0.5em minus 0.4em\relax {IEEE} Computer
  Society, pp. 929--934.

\bibitem{DBLP:conf/icann/HuangSHC19}
J.~Huang, H.~Shen, L.~Hou, and X.~Cheng, ``Signed graph attention networks,''
  in \emph{{ICANN} 2019}, ser. Lecture Notes in Computer Science, vol.
  11731.\hskip 1em plus 0.5em minus 0.4em\relax Springer, 2019, pp. 566--577.

\bibitem{DBLP:conf/aaai/LiTZC20}
Y.~Li, Y.~Tian, J.~Zhang, and Y.~Chang, ``Learning signed network embedding via
  graph attention,'' in \emph{{AAAI} 2020}.\hskip 1em plus 0.5em minus
  0.4em\relax {AAAI} Press, 2020, pp. 4772--4779.

\bibitem{berger1985certain}
J.~O. Berger, ``Certain standard loss functions,'' \emph{Statistical decision
  theory and Bayesian Analysis}, pp. 60--64, 1985.

\bibitem{morgan2015counterfactuals}
S.~L. Morgan and C.~Winship, \emph{Counterfactuals and causal inference}.\hskip
  1em plus 0.5em minus 0.4em\relax Cambridge University Press, 2015.

\bibitem{zhao2022learning}
T.~Zhao, G.~Liu, D.~Wang, W.~Yu, and M.~Jiang, ``Learning from counterfactual
  links for link prediction,'' in \emph{ICML}.\hskip 1em plus 0.5em minus
  0.4em\relax PMLR, 2022, pp. 26\,911--26\,926.

\bibitem{hartigan1979algorithm}
J.~A. Hartigan and M.~A. Wong, ``Algorithm as 136: A k-means clustering
  algorithm,'' \emph{Journal of the royal statistical society. series c
  (applied statistics)}, vol.~28, no.~1, pp. 100--108, 1979.

\bibitem{he2020lightgcn}
X.~He, K.~Deng, X.~Wang, Y.~Li, Y.~Zhang, and M.~Wang, ``Lightgcn: Simplifying
  and powering graph convolution network for recommendation,'' in \emph{SIGIR},
  2020, pp. 639--648.

\bibitem{DBLP:journals/pvldb/HuangLYC15}
X.~Huang, L.~V.~S. Lakshmanan, J.~X. Yu, and H.~Cheng, ``Approximate closest
  community search in networks,'' \emph{{VLDBJ}}, vol.~9, no.~4, pp. 276--287,
  2015.

\bibitem{mehlhorn1988faster}
K.~Mehlhorn, ``A faster approximation algorithm for the steiner problem in
  graphs,'' \emph{Information Processing Letters}, vol.~27, no.~3, pp.
  125--128, 1988.

\bibitem{WangC12}
J.~Wang and J.~Cheng, ``Truss decomposition in massive networks,''
  \emph{PVLDB}, vol.~5, no.~9, pp. 812--823, 2012.

\bibitem{read2009classifier}
J.~Read, B.~Pfahringer, G.~Holmes, and E.~Frank, ``Classifier chains for
  multi-label classification,'' in \emph{ECML-KDD}.\hskip 1em plus 0.5em minus
  0.4em\relax Springer, 2009, pp. 254--269.

\bibitem{wright1995logistic}
R.~E. Wright, ``Logistic regression.'' 1995.

\bibitem{suykens1999least}
J.~A. Suykens and J.~Vandewalle, ``Least squares support vector machine
  classifiers,'' \emph{Neural processing letters}, vol.~9, no.~3, pp. 293--300,
  1999.

\bibitem{bao2016intelligent}
Y.~Bao and X.~Jiang, ``An intelligent medicine recommender system framework,''
  in \emph{2016 IEEE 11Th conference on industrial electronics and applications
  (ICIEA)}.\hskip 1em plus 0.5em minus 0.4em\relax IEEE, 2016, pp. 1383--1388.

\bibitem{berg2017graph}
R.~v.~d. Berg, T.~N. Kipf, and M.~Welling, ``Graph convolutional matrix
  completion,'' \emph{arXiv preprint arXiv:1706.02263}, 2017.

\bibitem{chung2014empirical}
J.~Chung, C.~Gulcehre, K.~Cho, and Y.~Bengio, ``Empirical evaluation of gated
  recurrent neural networks on sequence modeling,'' \emph{arXiv preprint
  arXiv:1412.3555}, 2014.

\bibitem{jarvelin2002cumulated}
K.~J{\"a}rvelin and J.~Kek{\"a}l{\"a}inen, ``Cumulated gain-based evaluation of
  ir techniques,'' \emph{TOIS}, vol.~20, no.~4, pp. 422--446, 2002.

\bibitem{kingma2014adam}
D.~P. Kingma and J.~Ba, ``Adam: A method for stochastic optimization,''
  \emph{arXiv preprint arXiv:1412.6980}, 2014.

\bibitem{glorot2011deep}
X.~Glorot, A.~Bordes, and Y.~Bengio, ``Deep sparse rectifier neural networks,''
  in \emph{Proceedings of the fourteenth international conference on artificial
  intelligence and statistics}.\hskip 1em plus 0.5em minus 0.4em\relax JMLR
  Workshop and Conference Proceedings, 2011, pp. 315--323.

\bibitem{ioffe2015batch}
S.~Ioffe and C.~Szegedy, ``Batch normalization: Accelerating deep network
  training by reducing internal covariate shift,'' in \emph{ICML}.\hskip 1em
  plus 0.5em minus 0.4em\relax PMLR, 2015, pp. 448--456.

\bibitem{osheroff2007roadmap}
J.~A. Osheroff, J.~M. Teich, B.~Middleton, E.~B. Steen, A.~Wright, and D.~E.
  Detmer, ``A roadmap for national action on clinical decision support,''
  \emph{Journal of the American medical informatics association}, vol.~14,
  no.~2, pp. 141--145, 2007.

\bibitem{segundo2017improvement}
U.~Segundo, L.~Ald{\'a}miz-Echevarr{\'\i}a, J.~L{\'o}pez-Cuadrado,
  D.~Buenestado, F.~Andrade, T.~A. P{\'e}rez, R.~Barrena, E.~G.
  P{\'e}rez-Yarza, and J.~M. Pikatza, ``Improvement of newborn screening using
  a fuzzy inference system,'' \emph{Expert Systems with Applications}, vol.~78,
  pp. 301--318, 2017.

\bibitem{seol2017causality}
J.-W. Seol, W.~Yi, J.~Choi, and K.~S. Lee, ``Causality patterns and machine
  learning for the extraction of problem-action relations in discharge
  summaries,'' \emph{International journal of medical informatics}, vol.~98,
  pp. 1--12, 2017.

\bibitem{elkin2018artificial}
P.~L. Elkin, D.~R. Schlegel, M.~Anderson, J.~Komm, G.~Ficheur, and L.~Bisson,
  ``Artificial intelligence: bayesian versus heuristic method for diagnostic
  decision support,'' \emph{Applied clinical informatics}, vol.~9, no.~02, pp.
  432--439, 2018.

\bibitem{ma2016first}
S.~Ma, I.~R. Galatzer-Levy, X.~Wang, D.~Feny{\"o}, and A.~Y. Shalev, ``A first
  step towards a clinical decision support system for post-traumatic stress
  disorders,'' in \emph{AMIA Annual Symposium Proceedings}, vol. 2016.\hskip
  1em plus 0.5em minus 0.4em\relax American Medical Informatics Association,
  2016, p. 837.

\bibitem{li2020graph}
J.~Li, T.~Yu, D.-C. Juan, A.~Gopalan, H.~Cheng, and A.~Tomkins, ``Graph
  autoencoders with deconvolutional networks,'' \emph{arXiv preprint
  arXiv:2012.11898}, 2020.

\bibitem{baluja2008video}
S.~Baluja, R.~Seth, D.~Sivakumar, Y.~Jing, J.~Yagnik, S.~Kumar,
  D.~Ravichandran, and M.~Aly, ``Video suggestion and discovery for youtube:
  taking random walks through the view graph,'' in \emph{WWW}, 2008, pp.
  895--904.

\bibitem{eksombatchai2018pixie}
C.~Eksombatchai, P.~Jindal, J.~Z. Liu, Y.~Liu, R.~Sharma, C.~Sugnet, M.~Ulrich,
  and J.~Leskovec, ``Pixie: A system for recommending 3+ billion items to 200+
  million users in real-time,'' in \emph{WWW}, 2018, pp. 1775--1784.

\bibitem{wang2019unified}
Z.~Wang, H.~Liu, Y.~Du, Z.~Wu, and X.~Zhang, ``Unified embedding model over
  heterogeneous information network for personalized recommendation,'' in
  \emph{IJCAI}, 2019, pp. 3813--3819.

\bibitem{wang2018dkn}
H.~Wang, F.~Zhang, X.~Xie, and M.~Guo, ``Dkn: Deep knowledge-aware network for
  news recommendation,'' in \emph{WWW}, 2018, pp. 1835--1844.

\bibitem{li2021deconvolutional}
J.~Li, J.~Li, Y.~Liu, J.~Yu, Y.~Li, and H.~Cheng, ``Deconvolutional networks on
  graph data,'' in \emph{NeurIPS}, vol.~34, 2021, pp. 21\,019--21\,030.

\bibitem{chang2013know}
H.~chang, J.~Cai, and J.~Li, ``Knowledge graph completion with counterfactual
  augmentation,'' in \emph{WWW}, 2023.

\bibitem{feng2021should}
F.~Feng, W.~Huang, X.~He, X.~Xin, Q.~Wang, and T.-S. Chua, ``Should graph
  convolution trust neighbors? a simple causal inference method,'' in
  \emph{SIGIR}, 2021, pp. 1208--1218.

\bibitem{lin2021generative}
W.~Lin, H.~Lan, and B.~Li, ``Generative causal explanations for graph neural
  networks,'' in \emph{ICML}.\hskip 1em plus 0.5em minus 0.4em\relax PMLR,
  2021, pp. 6666--6679.

\bibitem{bevilacqua2021size}
B.~Bevilacqua, Y.~Zhou, and B.~Ribeiro, ``Size-invariant graph representations
  for graph classification extrapolations,'' in \emph{ICML}.\hskip 1em plus
  0.5em minus 0.4em\relax PMLR, 2021, pp. 837--851.

\bibitem{yu2022raneddi}
H.~Yu, W.~Dong, and J.~Shi, ``Raneddi: Relation-aware network embedding for
  drug-drug interaction prediction,'' \emph{Information Sciences}, vol. 582,
  pp. 167--180, 2022.

\bibitem{lin2015learning}
Y.~Lin, Z.~Liu, M.~Sun, Y.~Liu, and X.~Zhu, ``Learning entity and relation
  embeddings for knowledge graph completion,'' in \emph{AAAI}, 2015.

\bibitem{zitnik2016collective}
M.~Zitnik and B.~Zupan, ``Collective pairwise classification for multi-way
  analysis of disease and drug data,'' in \emph{Biocomputing 2016: Proceedings
  of the Pacific Symposium}.\hskip 1em plus 0.5em minus 0.4em\relax World
  Scientific, 2016, pp. 81--92.

\bibitem{wang2014knowledge}
Z.~Wang, J.~Zhang, J.~Feng, and Z.~Chen, ``Knowledge graph embedding by
  translating on hyperplanes,'' in \emph{AAAI}, vol.~28, no.~1, 2014.

\bibitem{wu2020comprehensive}
Z.~Wu, S.~Pan, F.~Chen, G.~Long, C.~Zhang, and S.~Y. Philip, ``A comprehensive
  survey on graph neural networks,'' \emph{TNNLS}, vol.~32, no.~1, pp. 4--24,
  2020.

\bibitem{feng2020dpddi}
Y.-H. Feng, S.-W. Zhang, and J.-Y. Shi, ``Dpddi: a deep predictor for drug-drug
  interactions,'' \emph{BMC bioinformatics}, vol.~21, no.~1, pp. 1--15, 2020.

\bibitem{wang2020gognn}
H.~Wang, D.~Lian, Y.~Zhang, L.~Qin, and X.~Lin, ``Gognn: Graph of graphs neural
  network for predicting structured entity interactions,'' \emph{arXiv preprint
  arXiv:2005.05537}, 2020.

\end{thebibliography}

\end{document}